\newcommand{\CIMIPublicVersion}{}
\newcommand{\CIMIIncludeSupplement}{}
\documentclass[letterpaper,10pt,conference]{ieeeconf}

\IEEEoverridecommandlockouts
\usepackage{booktabs}
\usepackage{multirow}
\usepackage{capt-of}
\usepackage{tabularx}
\usepackage{array}
\usepackage{amsmath}
\usepackage{amssymb}
\usepackage{url}
\usepackage{cite}
\ifdefined\CIMIIncludeSupplement
  \usepackage{chapterbib}
  \usepackage{balance}
  \DeclareMathSizes{6.7}{6.7}{5}{5}
\fi

\usepackage{xcolor}
\usepackage{colortbl}
\usepackage{tikz}
\usepackage{placeins}
\usepackage{wrapfig}
\usepackage{needspace}
\usetikzlibrary{arrows.meta,positioning}
\ifdefined\CIMIIncludeSupplement
  \usetikzlibrary{shadows}
\fi
\definecolor{urlblue}{HTML}{006EE7}
\makeatletter
\let\NAT@parse\undefined
\makeatother
\usepackage[colorlinks=true,linkcolor=urlblue,citecolor=urlblue,
  urlcolor=urlblue,bookmarks=false]{hyperref}
\ifdefined\CIMIPublicVersion\else
  \let\url\nolinkurl
\fi
\newcommand{\stageref}[1]{\hyperref[sec:stage#1]{\textbf{#1}}}
\newcommand{\suppref}[1]{%
  \ifdefined\CIMIIncludeSupplement
    Supp. Sec.~\ref{#1}%
  \else
    the supplementary material%
  \fi}
\makeatletter
\long\def\@makecaption#1#2{%
  \ifx\@captype\@IEEEtablestring
    \begingroup
      \footnotesize
      \parbox{\hsize}{\raggedright #1: #2\par}%
    \endgroup
    \@IEEEtablecaptionsepspace
  \else
    \@IEEEfigurecaptionsepspace
    \setbox\@tempboxa\hbox{\footnotesize #1.~~ #2}%
    \ifdim\wd\@tempboxa>\hsize
      \setbox\@tempboxa\hbox{\footnotesize #1.~~ }%
      \parbox[t]{\hsize}{\footnotesize\noindent\unhbox\@tempboxa#2}%
    \else
      \ifcenterfigcaptions
        \hbox to\hsize{\footnotesize\hfil\box\@tempboxa\hfil}%
      \else
        \hbox to\hsize{\footnotesize\box\@tempboxa\hfil}%
      \fi
    \fi
  \fi
}
\makeatother
\definecolor{cimiblue}{HTML}{DCEAF7}
\definecolor{cimigreen}{HTML}{E2F1E8}
\definecolor{cimiorange}{HTML}{F9E8D4}
\definecolor{cimipurple}{HTML}{EAE3F5}
\definecolor{tableyes}{HTML}{18B94E}
\definecolor{tableno}{HTML}{FF3333}
\newcommand{\tableyes}{\textcolor{tableyes}{\checkmark}}
\newcommand{\tableno}{\textcolor{tableno}{\times}}
\newcolumntype{P}[1]{>{\raggedright\arraybackslash}p{#1}}
\newcolumntype{C}[1]{>{\centering\arraybackslash}m{#1}}
\newcolumntype{Y}{>{\raggedright\arraybackslash}X}
\title{\LARGE \bf
Contact Imitation Interface:\\
Transferring Dexterous Skills by Showing How Contact Unfolds
}

\ifdefined\CIMIPublicVersion
\newcommand{\CIMIProjectURL}{https://nus-lins-lab.github.io/CIMIWeb/}
\IEEEaftertitletext{\vspace{-\baselineskip}}
\author{%
  Zhixuan Xu$^{1,*}$,\quad Yichen Li$^{1,2,*}$,\quad
  Bingchen Wu$^{1,2}$,\quad Rui Wang$^{1,2}$\\[0.2ex]
  Yiwen Hou$^{1}$,\quad Zixuan Liu$^{1}$,\quad
  Tianyu Qiu$^{2,3}$,\quad Lin Shao$^{1,2,\dagger}$\\[0.4ex]
  {\small $^{1}$National University of Singapore\quad $^{2}$RoboScience\quad
  $^{3}$South China University of Technology}\\[0.2ex]
  {\small $^{*}$Equal contribution\quad $^{\dagger}$Corresponding author}\\[0.4ex]
  {\small\bfseries Project page: \href{\CIMIProjectURL}{\texttt{nus-lins-lab.github.io/CIMIWeb/}}}%
}
\hypersetup{%
  pdftitle={Contact Imitation Interface: Transferring Dexterous Skills by Showing How Contact Unfolds},
  pdfauthor={Zhixuan Xu, Yichen Li, Bingchen Wu, Rui Wang, Yiwen Hou, Zixuan Liu, Tianyu Qiu, Lin Shao}%
}

\else
  \author{Anonymous Authors}
\fi

\ifdefined\CIMIIncludeSupplement
  \let\CIMIMakeTitle\maketitle
  \makeatletter
  \let\CIMIInternalMakeTitle\@maketitle
  \makeatother
\fi

\begin{document}

\ifdefined\CIMIIncludeSupplement
  \maketitle
\thispagestyle{empty}
\pagestyle{empty}

\begin{abstract}
Robot-free human demonstrations offer a promising route to scaling dexterous
manipulation data. The key challenge is to preserve the demonstrator's natural
dexterity while producing contact-grounded observations and actions that
transfer to robot hands. We present \textbf{Contact Imitation Interface (CIMI)},
a framework with contact-grounded demonstration and learning interfaces.
The demonstration interface combines shared contact geometry and fingertip RGB
views with contact-alignment-first retargeting and reprojection of wider scene
views. The learning interface introduces a hierarchical policy in which a
contact-surface planner predicts relative contact-surface trajectories and a
controller coordinates wrist and finger commands to improve contact alignment.
CIMI combines contact-grounded observations, retargeting, and action
representations, improving mean final-stage success
by at least 50 percentage points over baseline policies across six hand--task
pairs on LEAP and Wuji.
Hardware designs and software are publicly available.

\end{abstract}

\suppressfloats[t]
\begin{figure}[t]
  \centering
  \setlength{\abovecaptionskip}{2pt}
  \includegraphics[width=\columnwidth]{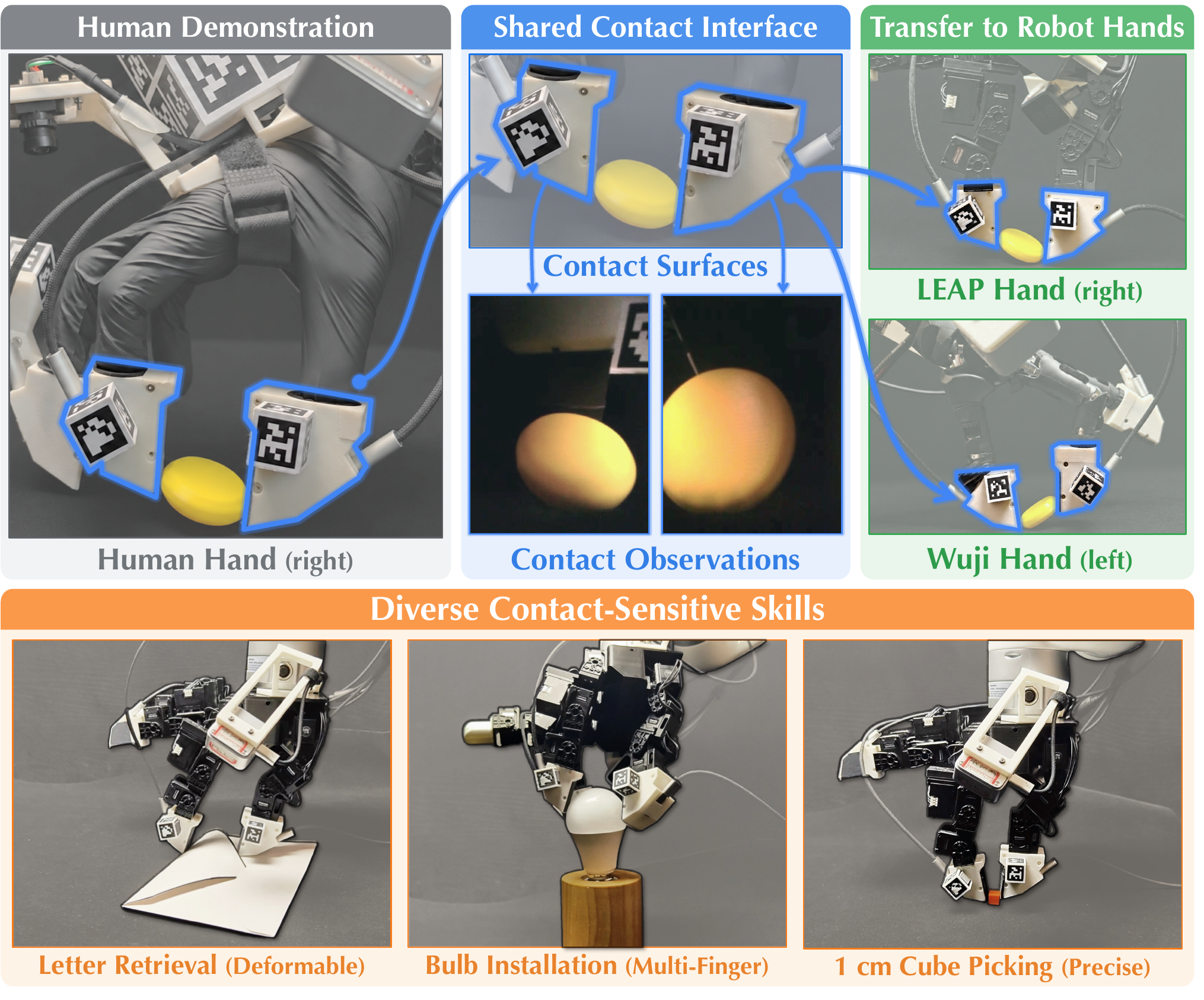}
  \caption{\textbf{CIMI transfers contact-sensitive skills from human demonstrations
  to robot hands.} Shared fingertip modules provide corresponding contact surfaces
  and camera views for human demonstration and execution across different robot hands. Bottom: tasks
  involving deformable objects, multi-finger coordination, and precise contact.}
  \label{fig:teaser}
\end{figure}

\section{Introduction}
\label{sec:introduction}

Low-cost teleoperation and portable hand-held grippers have enabled scalable
demonstration collection for parallel-jaw manipulation
\cite{zhao2023aloha,chi2024umi}. A human can fully control a
gripper's low-dimensional action space through its pose and opening, using
either the robot itself or a portable gripper to demonstrate manipulation.
Extending this approach to dexterous hands is especially challenging for
contact-sensitive tasks: differences in human and robot kinematics and
contact-surface geometry make it difficult to reproduce the demonstrated
sequence of contacts with the object.
This calls for a demonstration interface that preserves natural human dexterity
while capturing contact-grounded observations and motion that transfer to robot
hands.

Existing approaches anchor demonstrations in either the human or robot embodiment.
Human-centric interfaces prioritize natural manipulation with minimal
constraint. Video and motion capture recover human hand motion
\cite{wang2024dexcap,tao2025dexwild,zheng2026egoscale,qin2022dexmv}, while
instrumented gloves and wearable devices can additionally capture contact
geometry or tactile interaction
\cite{lin2026artglove,yin2025osmo,chen2026dexvitac,sathe2023fingertac}.
The recorded hand motion and sensory observations remain tied to the human
embodiment: retargeting may not
preserve the demonstrated contact evolution, and recorded observations may
differ from those available during robot execution. Robot-centric interfaces
instead establish compatibility through the collection hardware. Teleoperation
records robot-native observations and actions
\cite{zhao2023aloha,qin2023anyteleop,xu2026realdexumi}, while exoskeletons and
linkages mechanically map human motion to robot configurations
\cite{xu2025dexumi,fang2025dexop,du2025mile}. This compatibility often requires
robot-specific hardware or operator expertise and can restrict demonstration
motion. The resulting trade-off is between demonstrator freedom and robot-side
fidelity (Table~\ref{tab:interfaces}).

\begin{table*}[t]
  \setlength{\abovecaptionskip}{2pt}
  \centering
  \fontsize{8}{9}\selectfont
  \setlength{\tabcolsep}{2pt}
  \renewcommand{\arraystretch}{1.0}
  \renewcommand{\tabularxcolumn}[1]{m{#1}}
  \newcommand{\interfacerowrule}{\noalign{
    {\color{black!25}\hrule height 0.25pt}\vskip2pt}}
  \begin{tabularx}{\textwidth}{@{}C{70pt}C{28pt}C{54pt}C{58pt}C{80pt}C{70pt}>{\centering\arraybackslash}X C{40pt}@{}}
      \toprule
      \multirow{2}{*}[-2pt]{Interface} & \multirow{2}{*}[-2pt]{Focus}
        & \multicolumn{2}{c}{Demonstrator dexterity}
        & \multicolumn{3}{c}{Demonstration--execution interface}
        & \multirow{2}{40pt}[-2pt]{\centering Cross-hand\newline reuse} \\
      \cmidrule(lr){3-4}\cmidrule(l){5-7}
      & & \mbox{Motion freedom} & \mbox{Haptic feedback} & \mbox{Contact geometry}
        & \mbox{Contact observations} & \mbox{Action mapping} & \\
      \midrule
      \shortstack[c]{Bare-hand video\\\cite{qin2022dexmv,shaw2022videodex,
        chen2025vividex,kareer2024egomimic,paliwal2026doasido,
        hoque2026egodex,zheng2026egoscale,punamiya2026egoverse,
        chen2026videomanip}}
        & Human & Unconstrained & Direct touch & Mismatched human contact surfaces
        & None & Hand-pose retargeting
        & $\tableyes$ \\
      \interfacerowrule
      \shortstack[c]{Sensing glove\\\cite{song2025opentouch,yin2025osmo,
        chen2026dexvitac,lin2026artglove}}
        & Human & Near-free & Through glove & Glove contact surfaces; matching varies
        & Human/shared tactile (optional) & Glove-state retargeting
        & $\tableyes$ \\
      \interfacerowrule
      \shortstack[c]{Teleoperation\\\cite{handa2020dexpilot,qin2023anyteleop,
        iyer2024openteach,cheng2024opentelevision,ding2024bunnyvisionpro,
        lin2024hato,zhang2025doglove,gao2025glovity,koh2026dexmouse,
        chao2025exoviha,rayyan2026teledex,xu2026realdexumi}}
        & Robot & Controller/\allowbreak robot-limited & Remote/optional
        & Matched robot contact surfaces & Robot sensing (optional) & Native robot commands
        & $\tableyes/\tableno$ \\
      \interfacerowrule
      \shortstack[c]{Fingertip exoskeleton\\\cite{xu2025dexumi,si2025exostart,du2025mile}}
        & Robot & Exoskeleton-limited & Through contact surfaces
        & Matched robot contact surfaces & Matched tactile (optional)
        & Robot-aligned device state
        & $\tableno$ \\
      \interfacerowrule
      \shortstack[c]{Passive linkage\\\cite{fang2025dexop,zhu2026dexexo}}
        & Robot & Linkage-limited & Through linkage
        & Robot-aligned contact surfaces & Robot-aligned tactile (optional) & Direct joint transfer
        & $\tableno$ \\
      \interfacerowrule
      \textbf{CIMI (Ours)} & \textbf{Contact} & \textbf{Near-free}
        & \textbf{Through contact surfaces} & \textbf{Matched shared contact surfaces}
        & \textbf{Matched contact cameras}
        & \textbf{Contact-surface retargeting}
        & $\tableyes$ \\
      \bottomrule
  \end{tabularx}
  \par\vspace{4pt}
  \caption{Demonstration interfaces: dexterity, contact geometry and observations
  relative to execution, and cross-hand reuse. Mixed entries vary across cited systems.}
  \label{tab:interfaces}
\end{table*}

To address this trade-off, we introduce the \textbf{Contact Imitation Interface (CIMI)},
which combines contact-grounded demonstration and learning interfaces:

\textbf{1) We design a contact-grounded demonstration interface.}
Human and robot hands use fingertip modules with matching contact-surface
geometry and camera placement. Each module includes an RGB camera, termed a
\emph{contact camera}, that captures local fingertip--object interactions.
Cameras mounted at the wrist and thumb--index web space (\emph{context cameras})
provide wider scene views for manipulation.
Transferring demonstrations requires aligning contact motion and
visual observations between human and robot hands. However, differences in
hand kinematics can prevent simultaneous alignment of contact surfaces and
context-camera poses. We therefore propose contact-alignment-first retargeting
and context-view reprojection. By jointly optimizing wrist and finger motion,
retargeting prioritizes reproducing the demonstrated contact; we retain the
contact-camera images and reproject context views to the robot
viewpoints.

\textbf{2) We develop a contact-grounded learning interface.}
Policies commonly predict relative wrist motion and finger-joint commands
\cite{xu2025dexumi}. In contact-sensitive tasks with high-degree-of-freedom hands,
we observe that this representation can lead to poor wrist--finger coordination
and contact misalignment. We propose a hierarchical policy where a
contact-surface planner predicts relative contact-surface trajectories from visual
and proprioceptive inputs. A contact-surface controller converts these trajectories
into coordinated wrist and finger commands.

By centering both demonstration and learning on contact surfaces, CIMI enables
contact-sensitive dexterous skills to be learned from natural human
demonstrations and reproduced across different robot hands. Experiments on LEAP
and Wuji cover contact-sensitive tasks including grasping a 1\,cm cube, wedging
and lifting a flat coin, retrieving a letter, and screwing in a bulb. Across six
hand--task pairs, CIMI improves mean success by at least 50 percentage points
over baseline policies.
\ifdefined\CIMIIncludeSupplement
Videos are available at
\else
Supplementary materials and videos are available at
\fi
\ifdefined\CIMIPublicVersion
  \url{https://nus-lins-lab.github.io/CIMIWeb/}.
\else
  \url{https://contactimitationinterface.github.io/}.
\fi

\ifdefined\CIMIPublicVersion
\begin{figure*}[t]
  \centering
  \includegraphics[width=\textwidth]{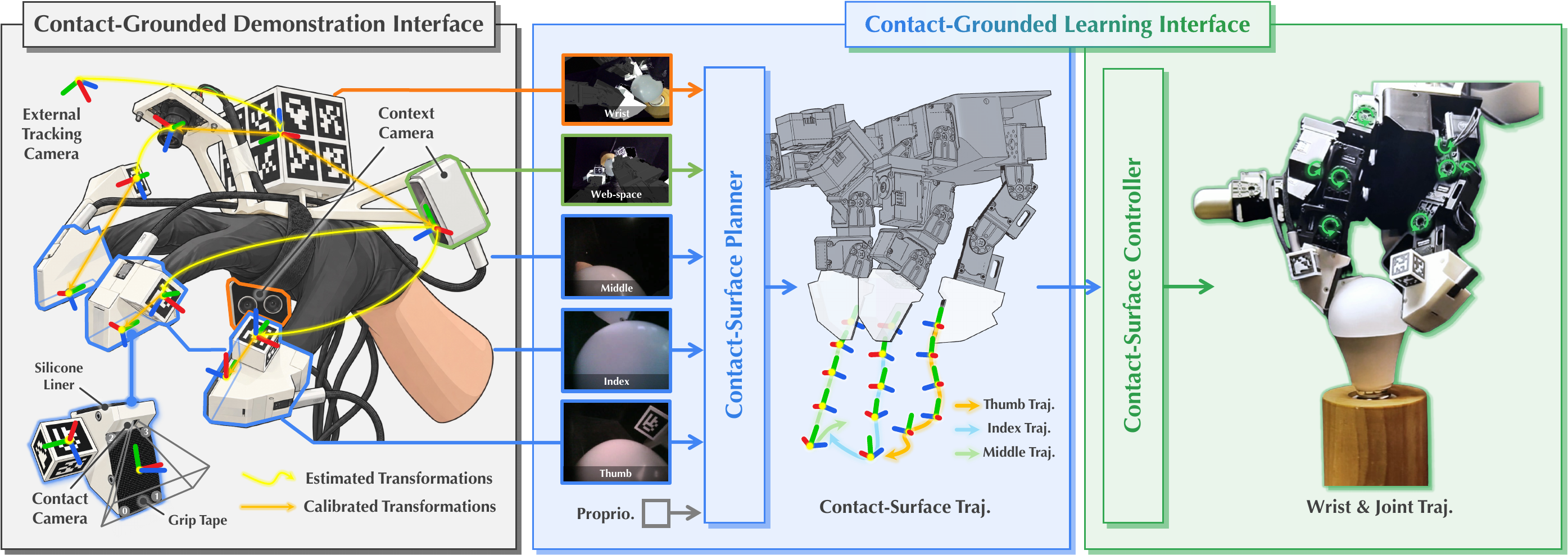}
  \caption{\textbf{Contact connects demonstration and learning in CIMI.} The demonstration
  interface captures contact-surface trajectories and observations, then constructs
  robot demonstrations through contact-alignment-first retargeting and
  context-view reprojection. The learning interface predicts relative contact-surface
  trajectories and realizes them through coordinated wrist and finger control.}
  \label{fig:method_overview}
\end{figure*}

\fi
\section{Related Work}
\label{sec:related-work}

\subsection{Demonstration Interfaces for Dexterous Manipulation}

Human-centric interfaces capture manipulation directly from the human hand.
Bare-hand approaches capture or reconstruct human-hand motion---and, when
available, object motion---from RGB or egocentric videos
\cite{qin2022dexmv,shaw2022videodex,kareer2024egomimic,
paliwal2026doasido,hoque2026egodex,zheng2026egoscale,
punamiya2026egoverse}. Motion-capture systems provide more reliable hand and
wrist trajectories \cite{wang2024dexcap,tao2025dexwild}, while instrumented
gloves additionally capture touch, tactile signals, or contact-surface
trajectories \cite{song2025opentouch,yin2025osmo,chen2026dexvitac,
lin2026artglove}. For example, DexCap~\cite{wang2024dexcap} combines hand
motion capture with scene point clouds and fingertip-position retargeting.
CIMI provides contact-grounded observations from matched fingertip cameras.
Its contact-alignment-first retargeting improves contact alignment and
task success over DexCap-style retargeting
(Table~\ref{tab:retargeting_ablation}).

Robot-centric interfaces instead collect demonstrations through
robot-compatible hardware. Teleoperation maps human input to an actuated robot
hand and directly records robot actions and observations
\cite{handa2020dexpilot,qin2023anyteleop,iyer2024openteach,
cheng2024opentelevision,ding2024bunnyvisionpro,lin2024hato,
yin2025geort,yin2025dexteritygen,zhang2025doglove,gao2025glovity,
koh2026dexmouse,chao2025exoviha,xu2026realdexumi,zhang2026fingervip}. Fingertip exoskeletons
constrain or measure human motion in a robot-oriented space
\cite{xu2025dexumi,si2025exostart}, while linkage-based exoskeletons and proxy
hands mechanically align human motion with robot kinematics
\cite{fang2025dexop,zhu2026dexexo}. For example, DexUMI~\cite{xu2025dexumi}
maps exoskeleton motion directly to robot joints, constraining demonstrations
to the target hand's kinematics: 1--2 active DoFs per finger on Inspire and
2--3 on XHand. CIMI supports four active DoFs per finger without imposing
robot-specific finger kinematics on the demonstrator. Its contact-grounded
planner and controller improve success over direct wrist--joint prediction
on our contact-sensitive tasks (Table~\ref{tab:action_representation}).

\subsection{Learning Dexterous Manipulation from Human Demonstrations}

We focus on learning robot policies from demonstrations captured through human
hand motion and sensing, rather than transferring fully specified hand--object
motion-capture trajectories \cite{li2025maniptrans,chen2024objectcentric}. The
key challenge is aligning human demonstrations with robot execution in both
action and observation spaces. Existing methods address this challenge through
explicit transformation, cross-embodiment policy learning, or both. For action
alignment, human wrist, hand, or fingertip motion is retargeted into robot joint
or end-effector commands \cite{qin2022dexmv,wang2024dexcap,xu2025dexumi,
chen2026dexvitac,zhang2026unidex,zheng2026egoscale}. For observation alignment,
demonstration videos can be transformed through human-hand segmentation,
background inpainting, and robot-hand compositing, or matched sensing modalities
can be used during both demonstration and execution
\cite{xu2025dexumi,yin2025osmo,
chen2026dexvitac}. Learning-based approaches use pretraining, co-training, or
representation alignment to ground human-derived representations in robot
sensing and control \cite{kareer2024egomimic,tao2025dexwild,
zheng2026egoscale,punamiya2026egoverse}.

CIMI uses matched fingertip contact surfaces as a shared cross-embodiment
interface: tracked contact-surface trajectories supervise robot actions, while local sensing
aligns policy observations, without imposing robot-specific hand kinematics on
the demonstrator. A contact-surface planner and controller use this shared
representation to separate contact-surface trajectory planning from wrist and
finger control.

\ifdefined\CIMIPublicVersion\else
\begin{figure*}[t]
  \centering
  \includegraphics[width=\textwidth]{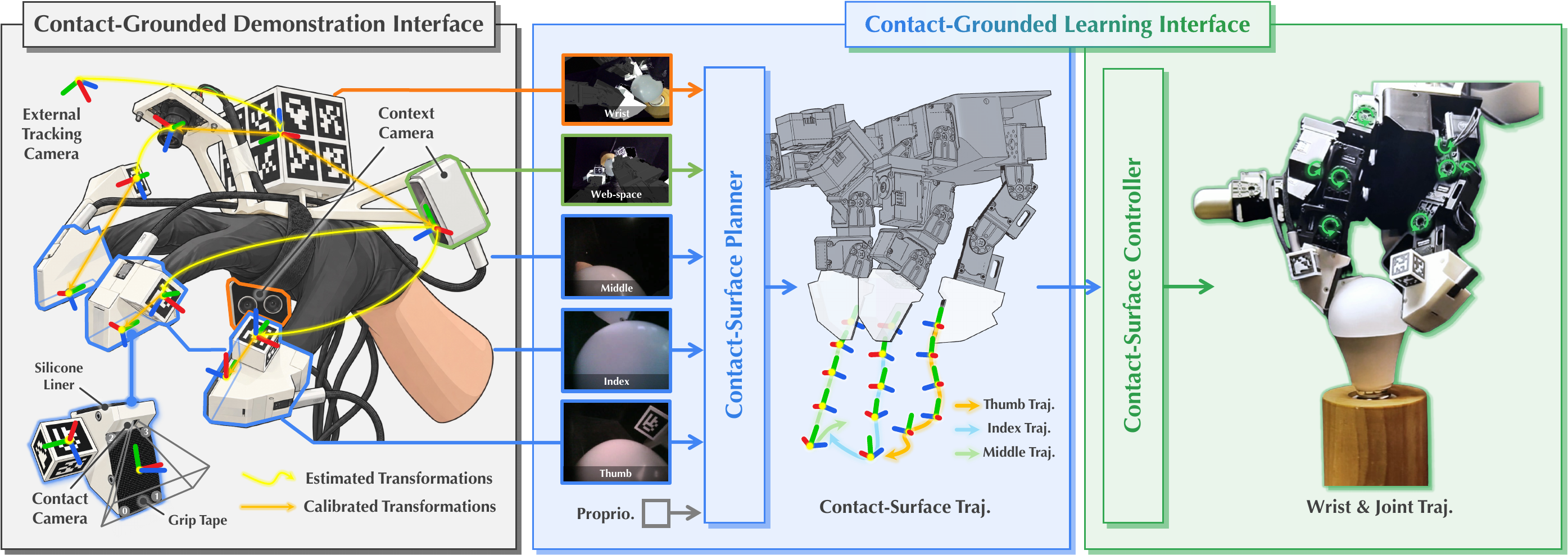}
  \caption{\textbf{Contact connects demonstration and learning in CIMI.} The demonstration
  interface captures contact-surface trajectories and observations, then constructs
  robot demonstrations through contact-alignment-first retargeting and
  context-view reprojection. The learning interface predicts relative contact-surface
  trajectories and realizes them through coordinated wrist and finger control.}
  \label{fig:method_overview}
\end{figure*}

\fi
\section{Contact Imitation Interface}
\label{sec:method}

\newcommand{\methodparagraph}[1]{%
  \par\addvspace{0.4\baselineskip}%
  \phantomsection%
  \noindent\textbf{#1.}\enspace\ignorespaces}

CIMI uses contact surfaces to connect human demonstration and
robot learning (Fig.~\ref{fig:method_overview}). The demonstration interface
transfers contact motion and visual observations to robot hands
(Section~\ref{sec:demo_interface}). The learning interface predicts relative
contact-surface trajectories and converts them into coordinated wrist and finger
commands (Section~\ref{sec:learning_interface}).

\subsection{Contact-Grounded Demonstration Interface}
\label{sec:demo_interface}
\label{sec:cimi_hardware}

The demonstration interface converts human demonstrations into contact-grounded
robot motion targets and observations. We achieve this through four design
choices. Shared fingertip hardware establishes matching contact-surface geometry
and camera placement (\stageref{D1}), while visual tracking recovers demonstrated
contact-surface trajectories (\stageref{D2}). Contact-alignment-first retargeting
jointly optimizes wrist and finger motion to reproduce these trajectories
(\stageref{D3}). We retain contact-camera views and reproject context-camera
views to match the retargeted motion (\stageref{D4}).

\methodparagraph{D1. Contact-Grounded Capture Hardware}
\label{sec:stageD1}%
\label{sec:contact_capture_hardware}
CIMI captures contact-surface trajectories and observations using cameras and
printed fiducials.
Human and robot fingertip modules share contact-surface geometry and camera
placement. Each RGB camera (\emph{contact camera}) views the local contact region
from the fingertip root, tangentially to the surface.
Two Gemini 305 stereo cameras view the wider scene
(\emph{context cameras}) from the wrist and thumb--index web space.
We record nine synchronized streams at 10\,Hz: three contact-camera views (thumb,
index, and middle fingers), four context views, and middle-finger and external
tracking views.
Hardware specifications, fabrication details, and a bill of materials are
provided in \suppref{sec:supp_hardware}.

\methodparagraph{D2. Visual Contact-Surface Tracking}
\label{sec:stageD2}%
\label{sec:motion_transfer}
\label{sec:surface_tracking}
We recover contact-surface trajectories by composing estimated poses and
calibrated transforms (Fig.~\ref{fig:method_overview}). We use AprilTag markers \cite{olson2011apriltag}
on AprilCube targets \cite{park2026aprilcube}, with DeepTag \cite{zhang2023deeptag}
providing dense grid keypoints. We jointly fit a target pose to keypoints across visible
tags in one multi-tag Perspective-$n$-Point (PnP) problem \cite{xu2026fingereye}.
Visible tags provide redundant constraints when other tags are occluded.
Frame definitions and estimator details are provided in \suppref{sec:supp_tracking}.

\methodparagraph{D3. Contact-Alignment-First Retargeting}
\label{sec:stageD3}%
\label{sec:surface_retargeting}
Our retargeting prioritizes contact-surface alignment over context-camera poses.
Fixing the wrist trajectory to align human and robot
context-camera poses (\emph{context-alignment-first}) leaves only finger joints
to reduce contact error (Fig.~\ref{fig:retargeting_priorities}a).
Following HoMMI's relaxation of camera-pose constraints~\cite{xu2026hommi},
we free the wrist to improve contact alignment and correct the resulting
viewpoint mismatch through reprojection
(Fig.~\ref{fig:retargeting_priorities}b; \stageref{D4}).
Table~\ref{tab:retargeting_ablation} evaluates this choice. Human contact-surface
poses and retargeted robot trajectories share the external tracking-camera frame.
We optimize wrist poses $\mathbf X_t$ and finger joints $\mathbf q_t$ to match
robot corners $\mathrm{FK}_{i,\ell}(\mathbf X_t,\mathbf q_t)$ to tracked human
corners $\mathbf p^H_{i,\ell,t}$, where $\mathrm{FK}$ denotes forward kinematics.
Matching four corners $\ell$ per finger $i$ (Fig.~\ref{fig:method_overview}, left)
aligns contact-surface position and orientation:
\begin{equation*}
\resizebox{\columnwidth}{!}{$\displaystyle
\{\mathbf X_t^*,\mathbf q_t^*\}_{t=1}^{T}
=\operatorname*{argmin}_{\{\mathbf X_t,\mathbf q_t\}_{t=1}^{T}}
\Bigl[\sum_{t,i,\ell}
\|\mathrm{FK}_{i,\ell}(\mathbf X_t,\mathbf q_t)-\mathbf p^H_{i,\ell,t}\|_2^2
+\mathcal R\Bigr].$}
\end{equation*}
The smoothness regularizer $\mathcal R$ penalizes changes in wrist translation,
rotation, and finger joints between valid neighboring frames.
We initialize the optimization by aligning human and robot
context-camera poses. We solve the retargeting optimization using
PyRoki~\cite{kim2025pyroki}, subject to the robot's finger-joint limits.
From the retargeted trajectory, we derive relative contact-surface trajectories to
supervise the planner (Section~\ref{sec:learning_interface}). The wrist and joint
trajectories also provide the basis for motion augmentation and controller
training. The retargeting objective and optimization settings are detailed in
\suppref{sec:supp_retargeting}.

\newlength{\methodpairheight}
\begin{figure*}[t]
  \setlength{\methodpairheight}{0.246\textwidth}
  \begin{minipage}[t]{0.43\textwidth}
  \vspace{0pt}
  \centering
  \includegraphics[height=\methodpairheight]{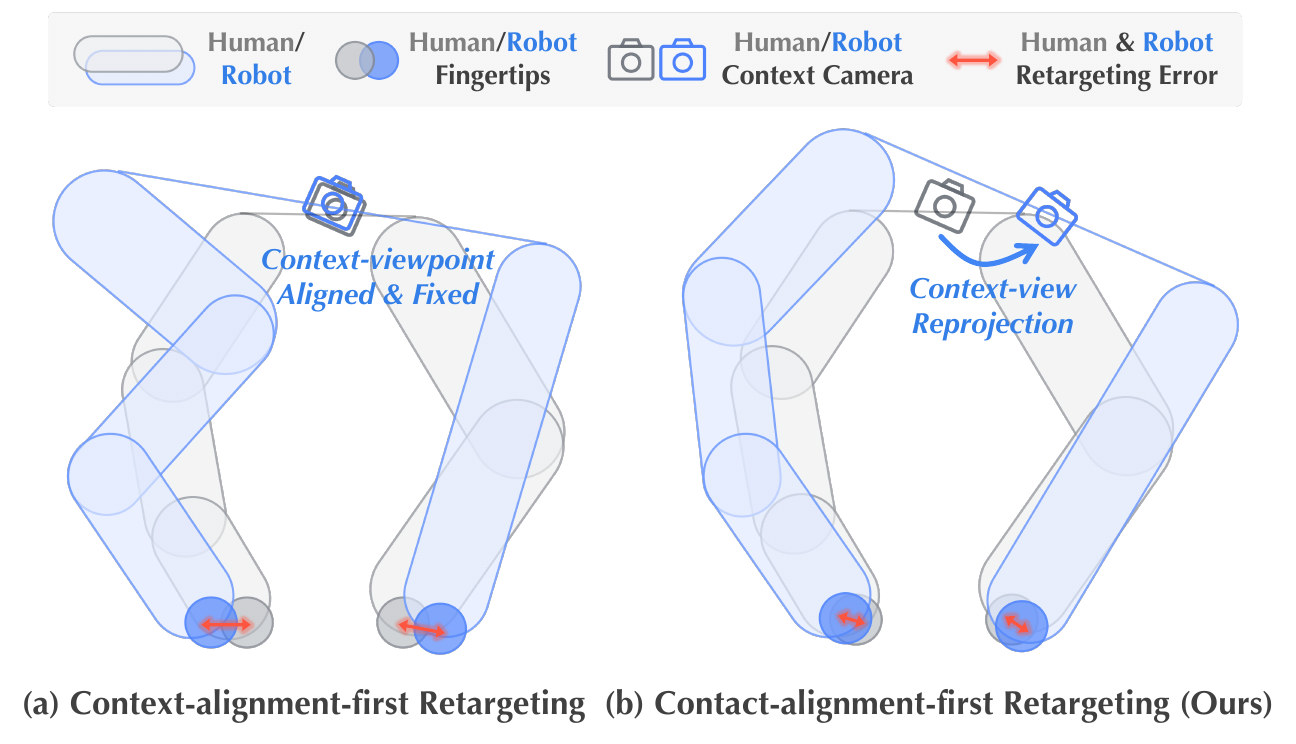}
  \caption{\textbf{Retargeting priorities.} (a) Context-alignment-first retargeting matches
  context-camera poses but can leave contact surfaces misaligned. (b) Contact-alignment-first
  retargeting adjusts the wrist to improve contact alignment;
  reprojection corrects the viewpoint mismatch. Gray: human;
  blue: robot.}
  \label{fig:retargeting_priorities}
  \end{minipage}\hfill
  \begin{minipage}[t]{0.555\textwidth}
  \vspace{0pt}
  \centering
  \includegraphics[height=\methodpairheight]{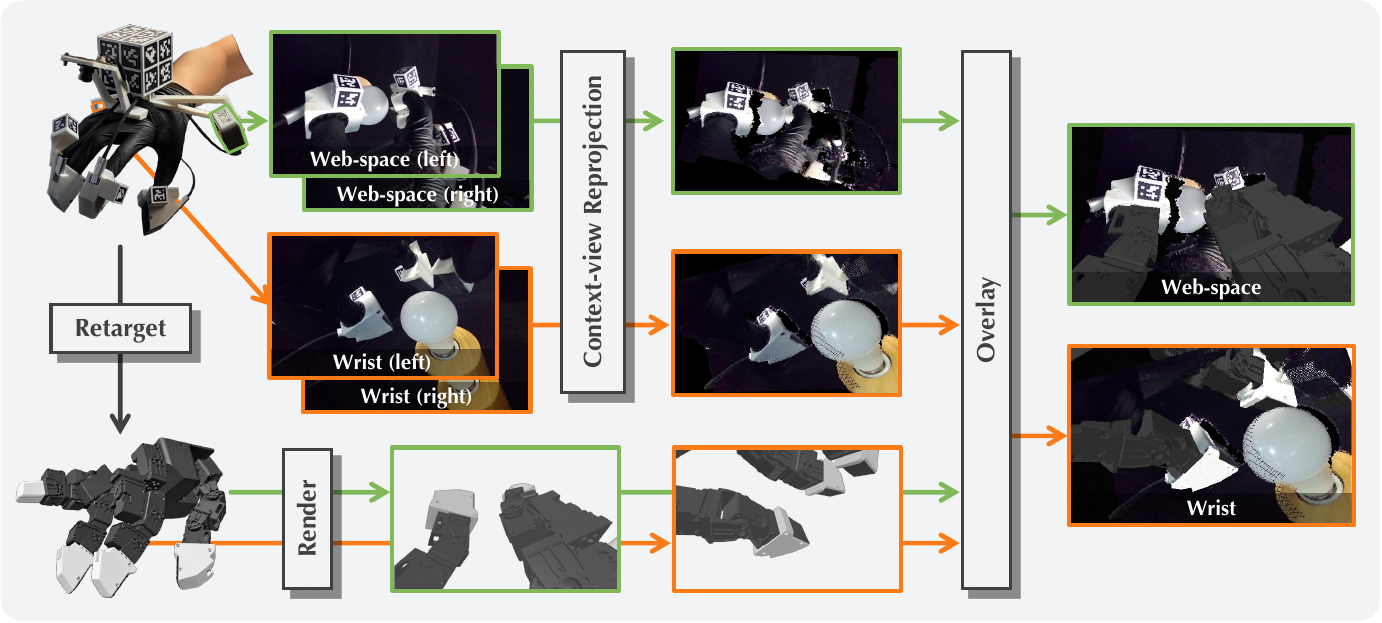}
  \caption{\textbf{Context-view reprojection and robot-geometry overlay.}
  Recorded web-space (green) and wrist (orange) stereo images are reprojected
  to camera poses determined by retargeting. Robot geometry is rendered at the
  retargeted configuration and overlaid on these views to produce RGB training
  observations from the robot perspective. This conversion is performed offline.}
  \label{fig:context_conversion}
  \end{minipage}
\end{figure*}

\methodparagraph{D4. Contact Observations and Context-View Reprojection}
\label{sec:stageD4}%
\label{sec:sensing_to_observations}
We construct demonstration observations to align with the contact and context views
available during robot execution. Shared contact-surface geometry and camera
placement give corresponding human and robot cameras the same nominal viewpoint
relative to the local contact surface, so we retain the recorded contact-camera
images. Context-camera viewpoints change with the retargeted wrist motion
(\stageref{D3}), so we convert their recorded views through reprojection and
robot-geometry overlay to match the robot perspective
(Fig.~\ref{fig:context_conversion}).
We optimize robot context-camera mounts using collected demonstrations
(details in \suppref{sec:supp_camera_mounts}).
For each stereo pair, we estimate depth with
Fast-FoundationStereo \cite{wen2026fastfoundationstereo} and back-project the RGB
images and estimated depth into a colored point cloud. We transform this cloud
into the robot camera frame determined by these mounts and the retargeted wrist and hand
configuration, reproject it onto the image plane, and composite it with the
robot geometry using nearest-depth visibility. The resulting RGB context views,
recorded contact-camera images, and retargeted wrist--hand states form
synchronized demonstration samples for policy training. Demonstration conversion
is performed offline. At execution, the policy receives contact RGB images and
preprocessed context RGB views; depth is used only for context-image preprocessing.

\subsection{Contact-Grounded Learning Interface}
\label{sec:learning_interface}
\label{sec:contact_imitation}

The learning interface separates contact-surface trajectory planning from
wrist--finger control to improve coordination (\stageref{O3}). Direct command
prediction \cite{xu2025dexumi} requires a visual policy to learn both contact
motion and its kinematic realization. We instead represent actions as relative
contact-surface trajectories (\stageref{L1}), train a visual planner to predict
them from demonstrations (\stageref{L2}), and train a separate controller to
convert them into wrist and finger commands using augmented synthetic motion
data (\stageref{L3}).

\methodparagraph{L1. Relative Contact-Surface Trajectories as Actions}
\label{sec:stageL1}%
\label{sec:relative_contact_trajectories}
The planner predicts how each contact surface should move to establish and
maintain contact with the object. We express this motion relative to the contact
surface's current pose, aligning actions with observations from the contact
camera on the same fingertip module.

Let $F$ be the number of controlled fingers, and let
$\mathrm{FK}_i(\mathbf q_t)$ denote the pose of contact surface $i$ in the wrist
frame. Its pose in a retargeted demonstration is
$\mathbf Y_{i,t}=\mathbf X_t^*\mathrm{FK}_i(\mathbf q_t^*)$.
For each contact surface, all future poses in a chunk use its frame at prediction
time $t$ as the reference, following the relative-trajectory convention
of \cite{chi2024umi}:
\begin{equation*}
\Delta\mathbf T_{i,t,h}=\mathbf Y_{i,t}^{-1}\mathbf Y_{i,t+h}.
\end{equation*}
Relative actions are invariant to fixed changes of the global frame's origin
and orientation. Each pose uses three translation coordinates and six rotation
coordinates (the rotation matrix's first two rows). Concatenating them across
$F$ fingers gives $\mathbf u_{t,h}\in\mathbb R^{9F}$.

\methodparagraph{L2. Contact-Surface Planner}
\label{sec:stageL2}%
\label{sec:contact_planner}
\label{sec:contact_policy_observations}
A frozen C-RADIOv3-B backbone \cite{heinrich2025radiov25} extracts a summary
token per contact or context view. View-specific projections and an MLP fuse
the tokens into
$\mathbf z_t^{\mathrm{vis}}$. The proprioceptive input
$\mathbf s_t^{\mathrm{prop}}$ encodes thumb-relative contact-surface poses,
exposing inter-finger geometry for contact coordination.
This input outperforms the tested alternatives in Table~\ref{tab:design_ablation}.
Given $\mathbf o_t=(\mathbf z_t^{\mathrm{vis}},\mathbf s_t^{\mathrm{prop}})$,
we train a flow-matching planner $\pi_\theta(\mathbf U_t\mid\mathbf o_t)$
\cite{lipman2023flowmatching} on demonstrated contact-surface trajectories.
It outputs an $H$-step relative contact-surface trajectory
$\widehat{\mathbf U}_t=[\widehat{\mathbf u}_{t,1},\ldots,
\widehat{\mathbf u}_{t,H}]$.
We use visual augmentation during planner training to reduce sensitivity to
reprojection artifacts, the appearance gap between rendered and real robot
images, and photometric differences between demonstration and robot cameras.
\ifdefined\CIMIIncludeSupplement
Training details are in \suppref{sec:supp_planner}.
\else
The supplement provides training details.
\fi

\ifdefined\CIMIPublicVersion
\definecolor{rolloutorange}{RGB}{255,128,0}
\definecolor{rolloutblue}{RGB}{0,128,255}
\definecolor{rolloutyellow}{RGB}{255,204,0}
\definecolor{rolloutleap}{RGB}{233,245,235}
\definecolor{rolloutwuji}{RGB}{231,239,255}
\definecolor{rolloutfailure}{RGB}{251,231,229}
\DeclareRobustCommand{\rolloutgroupkey}[1]{%
  \tikz[baseline=0.15ex]{\filldraw[fill=#1,draw=#1!65!black,line width=0.3pt]
    (0,0) rectangle (0.9em,0.65em);}}
\DeclareRobustCommand{\rolloutviewkey}{%
  \tikz[baseline=0.15ex]{\draw[rolloutorange,line width=0.8pt]
    (0,0) rectangle (0.9em,0.65em);}}
\DeclareRobustCommand{\rolloutlinekey}[1]{%
  \tikz[baseline=-0.5ex]{\draw[#1,line width=1.2pt]
    (0,0) -- (0.9em,0);}}
\begin{figure*}[t]
  \centering
  \includegraphics[width=\textwidth]{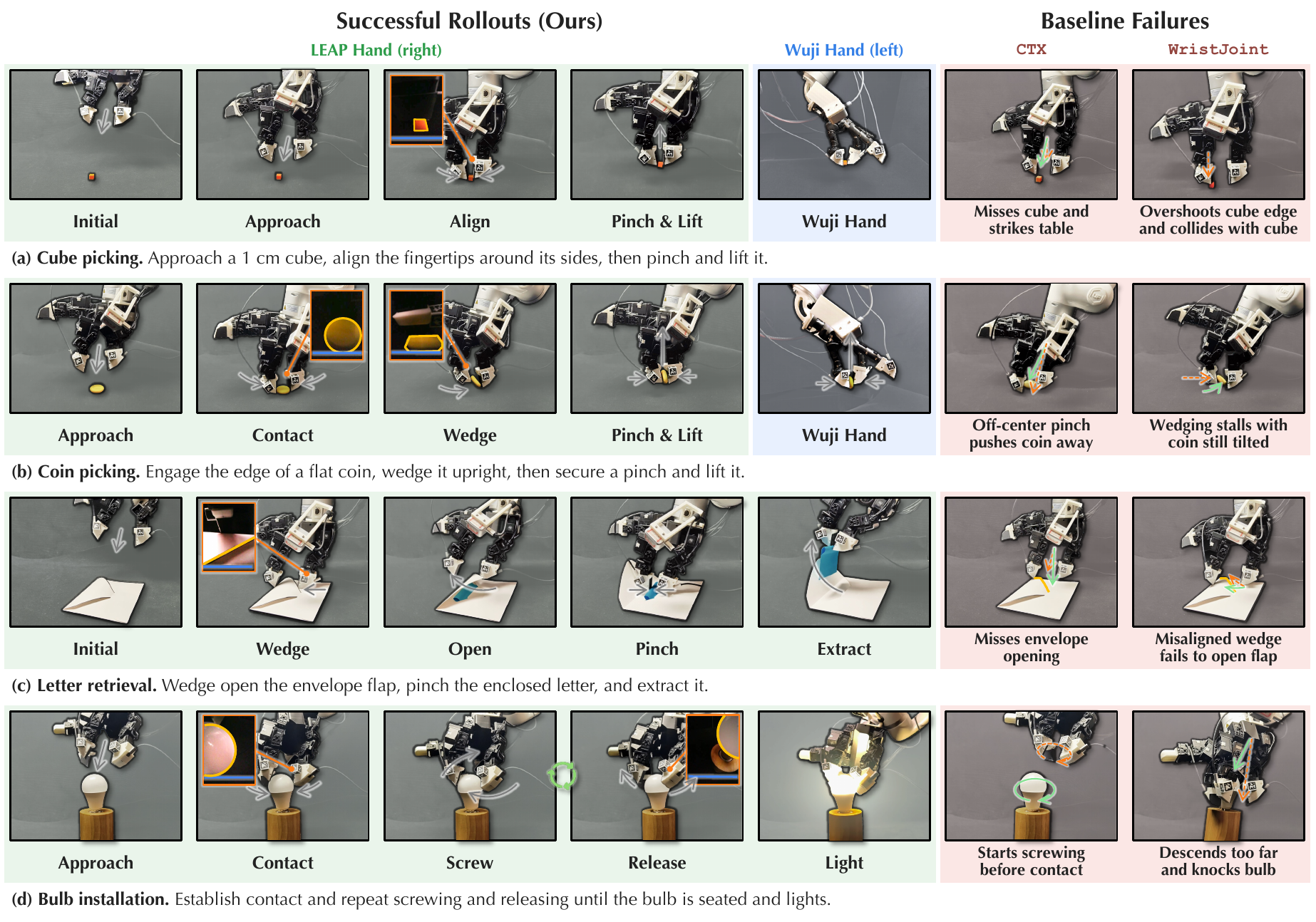}
  \caption{\textbf{Real-world rollouts and failure cases.}
  Orange-bordered insets (\rolloutviewkey) show contact-camera views;
  within each inset, blue lines (\rolloutlinekey{rolloutblue}) mark the contact surface
  and yellow outlines (\rolloutlinekey{rolloutyellow}) mark object boundaries.
  Green (\rolloutgroupkey{rolloutleap}) and blue (\rolloutgroupkey{rolloutwuji})
  backgrounds show successful LEAP and Wuji rollouts, respectively;
  red (\rolloutgroupkey{rolloutfailure}) marks \textbf{\texttt{CTX}} and \textbf{\texttt{WristJoint}} failures.}
  \label{fig:tasks_summary}
\end{figure*}

\begin{table*}[t]
  \setlength{\abovecaptionskip}{2pt}
  \centering
  \footnotesize
  \setlength{\tabcolsep}{2pt}
  \renewcommand{\arraystretch}{1.15}
  \begin{tabular*}{\textwidth}{@{\extracolsep{\fill}}ll*{8}{c}@{\hspace{8pt}}*{4}{c}@{\hspace{6pt}}c@{}}
    \toprule
    \multirow{3}{*}[-1ex]{Policy} & \multirow{3}{*}[-1ex]{Visual}
      & \multicolumn{8}{c}{\textbf{LEAP Hand}}
      & \multicolumn{4}{c}{\textbf{Wuji Hand}}
      & \multirow{3}{*}[-1ex]{\textbf{Mean}} \\
    \cmidrule(lr){3-10}
    \cmidrule(lr){11-14}
    & & \multicolumn{2}{c}{Cube}
      & \multicolumn{2}{c}{Coin}
      & \multicolumn{2}{c}{Letter}
      & \multicolumn{2}{c}{Bulb}
      & \multicolumn{2}{c}{Cube}
      & \multicolumn{2}{c}{Coin} & \\
    \cmidrule(lr){3-4}
    \cmidrule(lr){5-6}
    \cmidrule(lr){7-8}
    \cmidrule(lr){9-10}
    \cmidrule(lr){11-12}
    \cmidrule(lr){13-14}
    &
      & Align & Lift & Wedge & Lift & Open & Lift & Screw & Light
      & Align & Lift & Wedge & Lift & \\
    \midrule
    \textbf{\texttt{WristJoint}} & \textbf{\texttt{CTX}}
      & 0 & 0 & 0 & 0 & 0 & 0 & 35 & 5 & 0 & 0 & 0 & 0 & 0.8 \\
    \textbf{\texttt{ContactSurface}} & \textbf{\texttt{CTX}}
      & 0 & 0 & 0 & 0 & 0 & 0 & 50 & 10 & 0 & 0 & 0 & 0 & 1.7 \\
    \midrule
    \textbf{\texttt{WristJoint}} & \textbf{\texttt{CTX+CC}}
      & 35 & 20 & 50 & 35 & 25 & 15 & 85 & 70 & 30 & 15 & 40 & 30 & 30.8 \\
    \textbf{\texttt{ContactSurface}} & \textbf{\texttt{CTX+CC}}
      & \textbf{85} & \textbf{75} & \textbf{95} & \textbf{85} & \textbf{90} & \textbf{80} & \textbf{95} & \textbf{90} & \textbf{75} & \textbf{70} & \textbf{90} & \textbf{85} & \textbf{80.8} \\
    \bottomrule
  \end{tabular*}
  \par\vspace{4pt}
  \caption{Policy and observation comparison on LEAP and Wuji hands. SR in
  \%; 20 rollouts per method, hand, and task. Mean SR averages final-stage SR
  equally over all six hand--task pairs.}
  \label{tab:action_representation}
\end{table*}

\fi

\methodparagraph{L3. Contact-Surface Controller}
\label{sec:stageL3}%
\label{sec:contact_controller}
The MLP controller $C_\phi$ maps current joint values $\mathbf q_t$ and the predicted
contact-surface trajectory $\widehat{\mathbf U}_t$ to future wrist poses
$\widehat{\Delta\mathbf X}_{t,h}$ relative to the current wrist pose and absolute
finger-joint targets $\widehat{\mathbf q}_{t,h}$:
\begin{equation*}
\left[(\widehat{\Delta\mathbf X}_{t,h},\widehat{\mathbf q}_{t,h})\right]_{h=1}^{H}
=C_\phi(\mathbf q_t,\widehat{\mathbf U}_t).
\end{equation*}

To cover predicted contact-surface trajectories beyond recorded demonstrations,
we augment retargeted trajectories from \stageref{D3} with smooth wrist and
joint perturbations. Forward kinematics provides paired contact-surface
trajectories and wrist--joint commands.
We supervise the controller with wrist--joint command regression
and contact-surface alignment measured through forward kinematics.
Input construction, augmentation, and loss details are provided in
\suppref{sec:supp_controller}.

At execution, the controller converts the trajectory into commands in one forward
pass. We use $H=16$ and execute an 8-step prefix, then replan from the latest
observation. We reuse the controller across tasks with the same robot hand and
controlled fingers.

\section{Experiments}
\label{sec:experiments}

\newcommand{\experimentparagraph}[1]{%
  \par\addvspace{0.35\baselineskip}%
  \phantomsection%
  \noindent\textbf{#1.}\enspace\ignorespaces}

\ifdefined\CIMIPublicVersion\else
  
\fi

We evaluate whether CIMI's contact-grounded demonstration and learning
interfaces enable robot hands to learn contact-sensitive dexterous manipulation
from human demonstrations.
The experiments answer six questions:
\textbf{Q1: Learning across robot hands.} Can CIMI enable different robot
hands to learn contact-sensitive skills from the same human demonstrations?
\textbf{Q2: Contact-grounded observations.} Do contact-grounded observations
improve manipulation performance over context vision alone?
\textbf{Q3: Contact-grounded policy.} Do hierarchical contact-surface
planning and control improve manipulation over direct wrist and joint prediction?
\textbf{Q4: Contact-surface alignment.} Is accurate contact-surface alignment
important for contact-sensitive manipulation?
\textbf{Q5: Collection efficiency.} How fast can CIMI collect demonstrations
compared with bare-hand demonstration and teleoperation?
\textbf{Q6: Design ablations.} Which design choices contribute most to reliable
manipulation with CIMI?

\subsection{Q1--Q3: Contact-Grounded Policy Evaluation}

\experimentparagraph{Tasks and Evaluation Protocol}
Fig.~\ref{fig:tasks_summary} shows LEAP and Wuji rollouts.
LEAP tasks comprise cube picking (146 training demos), coin picking (187 demos),
letter retrieval (183 demos), and bulb installation (103 demos); Wuji tasks
comprise cube picking (same 146 demos) and coin picking (same 187 demos).
These tasks test fingertip alignment, wedging, envelope opening, and repeated
bulb regrasping. We report stage-wise success rates (SR) over 20 rollouts per
method, hand, and task; mean SR averages final-stage success across the six
hand--task pairs (Table~\ref{tab:action_representation}).

\experimentparagraph{Compared Policies}
Table~\ref{tab:action_representation} crosses two visual inputs---\textbf{\texttt{CTX}}
(task-specific context-camera views) and \textbf{\texttt{CTX+CC}} (the same context
views plus contact-camera images)---with two policies:
\textbf{\texttt{WristJoint}} (direct prediction of relative wrist motion
$(\mathbf X_t^*)^{-1}\mathbf X_{t+h}^*$ and joint changes
$\mathbf q_{t+h}^*-\mathbf q_t^*$, following DexUMI~\cite{xu2025dexumi}) and
\textbf{\texttt{ContactSurface}} (our contact-grounded policy).
Context views include wrist RGB and, depending on the task, web-space RGB;
they are matched between \textbf{\texttt{CTX}} and \textbf{\texttt{CTX+CC}}.
\ifdefined\CIMIIncludeSupplement
\suppref{sec:supp_policy_cameras} specifies the camera views for each task.
\else
The supplementary material specifies the camera views for each task.
\fi
All variants for a given hand use the same demonstrations and retargeted
trajectories. Within each visual setting, policy comparisons match visual and
proprioceptive inputs, flow-model width, and planner training budget. For cube and coin picking,
we retarget the same human recordings to each hand and train separate policies.

\experimentparagraph{O1: Human demonstrations support learning on both hands}
CIMI achieves 80.8\% mean final-stage SR across all six hand--task pairs
with \textbf{\texttt{ContactSurface}} and \textbf{\texttt{CTX+CC}}
(Table~\ref{tab:action_representation}). For the two tasks shared across hands,
cube lifting succeeds in 75\% of LEAP rollouts and 70\% of Wuji rollouts,
while coin lifting reaches 85\% on both. LEAP also achieves 80\% on letter
retrieval and 90\% on bulb installation. The successful rollouts in
Fig.~\ref{fig:tasks_summary} show policies learned through CIMI reproducing
demonstrated contact sequences on different robot hands.

\experimentparagraph{O2: Contact-camera feedback improves contact establishment}
Adding contact-camera views raises mean final-stage SR from 0.8\% to
30.8\% for \textbf{\texttt{WristJoint}} and from 1.7\% to 80.8\% for
\textbf{\texttt{ContactSurface}} (Table~\ref{tab:action_representation}).
With context vision alone, both policies score 0\% even at the first stage of
cube, coin, and letter tasks. The context-only examples in
Fig.~\ref{fig:tasks_summary} show errors at contact establishment: missing the
cube and striking the table, pushing the coin away with an off-center pinch,
missing the envelope opening, and starting to screw before contacting the bulb.
These examples illustrate the local placement and engagement cues that
contact-camera views provide beyond context vision.

\experimentparagraph{O3: Contact-surface planning better uses contact feedback}
\label{sec:stageO3}%
\label{sec:hierarchy_experiments}
With \textbf{\texttt{CTX+CC}}, \textbf{\texttt{ContactSurface}} exceeds
\textbf{\texttt{WristJoint}} by 50.0 percentage points in mean final-stage SR
(80.8\% versus 30.8\%). Gains span all six hand--task pairs, ranging from
20 points on LEAP bulb to 65 on LEAP letter, with 55-point gains on both Wuji
tasks. The \textbf{\texttt{WristJoint}} examples in
Fig.~\ref{fig:tasks_summary} show residual motion errors: overshooting the cube
edge, stalling while wedging the coin, misaligning with the envelope flap,
and descending too far onto the bulb. This pattern supports the design in
Section~\ref{sec:learning_interface}: plan motion in the contact-surface frame
and use a separate controller to coordinate wrist and finger commands.
The comparison evaluates this combined design; Q6 examines individual input
and augmentation choices.

\subsection{Q4: Contact-Surface Alignment}
\experimentparagraph{Tasks and Evaluation Protocol}
We compare center and corner matching under context- and contact-alignment-first
retargeting on LEAP coin picking and bulb installation
(Table~\ref{tab:retargeting_ablation}). Center matching aligns the centroid of
each surface's four corners without constraining orientation.
\emph{Context-alignment-first} fixes the wrist to match context-camera poses;
\emph{contact-alignment-first} jointly optimizes the wrist and finger joints
to align contact surfaces. All variants use the same human recordings. We measure mean
Euclidean error (mm) between retargeted and demonstrated surface corners
(four per finger)
(\stageref{D3}). For each task, we retrain planners for all variants with
identical data splits, architectures, training budgets, and controllers,
then evaluate each in 20 rollouts.

\begin{table}[t]
  \setlength{\abovecaptionskip}{0pt}
  \centering
  \footnotesize
  \setlength{\tabcolsep}{1.25pt}
  \renewcommand{\arraystretch}{1.15}
  \begin{tabularx}{\columnwidth}{*{8}{>{\centering\arraybackslash}X}}
    \toprule
    \multicolumn{2}{c}{Retargeting} & \multicolumn{3}{c}{Coin}
      & \multicolumn{3}{c}{Bulb} \\
    \cmidrule(lr){1-2}\cmidrule(lr){3-5}\cmidrule(l){6-8}
    \multirow{2}{*}{Target} & \multirow{2}{*}{Priority}
      & \multirow{2}{*}{\shortstack{Error\\(mm)$\downarrow$}}
      & \multicolumn{2}{c}{SR (\%)$\uparrow$}
      & \multirow{2}{*}{\shortstack{Error\\(mm)$\downarrow$}}
      & \multicolumn{2}{c}{SR (\%)$\uparrow$} \\
    \cmidrule(lr){4-5}\cmidrule(l){7-8}
      & & & Wedge & Lift & & Screw & Light \\
    \midrule
    \rowcolor{black!5}
    Centers & Context & 15.37 & 0 & 0 & 7.85 & 10 & 0 \\
    Corners & Context & 14.67 & 0 & 0 & 5.72 & 25 & 0 \\
    Centers & Contact & 9.31 & 0 & 0 & 4.36 & 50 & 20 \\
    \rowcolor{urlblue!8}
    \textbf{Corners} & \textbf{Contact} & \textbf{0.51} & \textbf{95} & \textbf{85} & \textbf{1.09} & \textbf{95} & \textbf{90} \\
    \bottomrule
  \end{tabularx}
  \par\vspace{2pt}
  \caption{Retargeting ablation on LEAP. Priority denotes context- or
  contact-alignment-first. Gray: DexCap-style~\cite{wang2024dexcap}; blue: ours.
  Error is the mean contact-corner distance over valid frames.
  SR uses 20 rollouts per variant and task; our SRs are from
  Table~\ref{tab:action_representation}.}
  \label{tab:retargeting_ablation}
\end{table}

\experimentparagraph{O4: Contact priority and corner matching jointly improve contact transfer}
All three alternative retargeting variants yield 0\% Coin Wedge SR but
10--50\% Bulb Screw SR (Table~\ref{tab:retargeting_ablation}). This contrast may reflect
greater tolerance to contact error during bulb engagement than during
coin-edge wedging. Completing bulb installation remains sensitive to alignment:
both context-alignment-first variants have 0\% Light SR. With contact priority, replacing
center matching with corner matching reduces bulb contact-corner error from 4.36 to
1.09\,mm and raises Light SR from 20\% to 90\%.

\Needspace{16\baselineskip}
\subsection{Q5: Collection Efficiency}
\experimentparagraph{Tasks and Evaluation Protocol}\par
\begin{wrapfigure}[12]{r}{0.51\columnwidth}
  \vspace{-8pt}
  \centering
  \setlength{\abovecaptionskip}{1pt}
  \setlength{\belowcaptionskip}{0pt}
  \includegraphics[width=\linewidth]{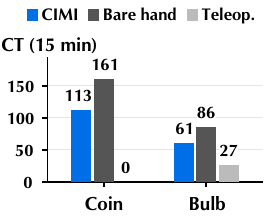}\par
  \caption{\textbf{15-minute collection throughput on coin picking and
  bulb installation.}}
  \label{fig:collection_efficiency}
\end{wrapfigure}
We compare CIMI, bare hands, and teleoperation by counting successful
demonstrations collected by the same operator in 15 minutes
(Fig.~\ref{fig:collection_efficiency}). We evaluate this collection throughput
(CT) on coin picking and bulb installation, using the same task success
criteria across interfaces. Teleoperation combines Manus hand tracking, Vive
wrist tracking, and GeoRT~\cite{yin2025geort} hand retargeting. The timed
protocol includes failed attempts, resets, and object randomization.

\experimentparagraph{O5: CIMI supports efficient collection of transferable demonstrations}
CIMI retains 70\% of bare-hand throughput on coin picking and 71\% on bulb
installation (Fig.~\ref{fig:collection_efficiency}). It achieves $2.3\times$
teleoperation throughput on bulb, while teleoperation yields no successful
coin demonstrations. Together with Table~\ref{tab:action_representation}, these
results support efficient demonstration collection for contact-sensitive robot
learning.

\subsection{Q6: Design Ablations}
\experimentparagraph{Tasks and Evaluation Protocol}
We ablate context-view conversion, planner proprioception, and visual and motion
augmentation on LEAP coin picking and bulb installation
(\mbox{Table~\ref{tab:design_ablation}}). Each variant changes the listed components
of \textbf{\texttt{ContactSurface}} with \textbf{\texttt{CTX+CC}}, using the same
stage criteria and 20-rollout evaluation protocol.

\begin{table}[t]
  \setlength{\abovecaptionskip}{0pt}
  \centering
  \footnotesize
  \setlength{\tabcolsep}{2pt}
  \renewcommand{\arraystretch}{1.15}
  \begin{tabularx}{\columnwidth}{>{\raggedright\arraybackslash}X*{5}{c}}
    \toprule
    \multirow{2}{*}[-0.5ex]{Variant} & \multicolumn{2}{c}{Coin}
      & \multicolumn{2}{c}{Bulb}
      & \multirow{2}{*}[-0.5ex]{\shortstack{Mean SR\\(\%)$\uparrow$}} \\
    \cmidrule(lr){2-3}\cmidrule(lr){4-5}
      & Wedge & Lift & Screw & Light & \\
    \midrule
    \multicolumn{6}{l}{\textit{Context-view conversion}} \\
    No overlay & 90 & 80 & 75 & 45 & 62.5 \\
    No reprojection, no overlay & 75 & 50 & 55 & 10 & 30.0 \\
    \midrule
    \multicolumn{6}{l}{\textit{Planner proprioceptive input}} \\
    No planner proprioception~\cite{xu2025dexumi} & 90 & 85 & 10 & 0 & 42.5 \\
    Joint values~\cite{zhao2023aloha,wang2024dexcap} & 70 & 55 & 80 & 50 & 52.5 \\
    Relative wrist + joints~\cite{chi2024umi,xu2026hommi} & 60 & 40 & 30 & 5 & 22.5 \\
    \midrule
    \multicolumn{6}{l}{\textit{Augmentation}} \\
    No planner visual augmentation & 0 & 0 & 15 & 0 & 0.0 \\
    Task-only controller, no motion aug. & 90 & 70 & 80 & 35 & 52.5 \\
    \midrule
    \rowcolor{urlblue!8}
    \textbf{Ours} & \textbf{95} & \textbf{85} & \textbf{95} & \textbf{90} & \textbf{87.5} \\
    \bottomrule
  \end{tabularx}
  \par\vspace{2pt}
  \caption{Design ablations on LEAP. Stage-wise SR in \%; 20 rollouts per
  variant and task. All variants use our
  \textbf{\texttt{ContactSurface}} + \textbf{\texttt{CTX+CC}} framework
  with the listed changes. Citations denote related input styles adapted to CIMI.
  Our SRs are from Table~\ref{tab:action_representation}.
  Mean SR averages Coin Lift and Bulb Light.}
  \label{tab:design_ablation}
\end{table}

\experimentparagraph{O6: Context-view conversion improves contact control}
\label{sec:stageO6}%
Removing the robot-geometry
overlay while retaining reprojection lowers mean SR from 87.5\% to 62.5\%;
using the original human context views without either operation lowers it
further to 30.0\% (Table~\ref{tab:design_ablation}). In these rollouts, overall
task motion remains reasonable, but contact control becomes less precise,
despite unchanged contact-camera inputs. The observed errors suggest that
reducing differences in context-camera viewpoints and robot appearance helps
refine contact control.

\experimentparagraph{O7: Thumb-relative proprioception improves planner performance}
\label{sec:stageO7}%
Thumb-relative contact-surface poses yield 87.5\% mean SR, compared with
52.5\% for joint-value inputs (style of~\cite{zhao2023aloha,wang2024dexcap}).
With joint values, coin rollouts show slightly
looser fingertip contact, and Coin Lift SR
falls from 85\% to 55\%. Removing planner proprioception
(style of~\cite{xu2025dexumi}) preserves 85\%
Coin Lift SR but reduces Bulb Light SR from 90\% to 0\%. In bulb rollouts,
relative-pose errors accumulate over successive motions, and the middle finger
drifts away from the bulb. The relative proprioception variant
(style of~\cite{chi2024umi,xu2026hommi}) uses the wrist pose relative to
the previous frame and inter-frame joint changes. This variant yields 22.5\%
mean SR and can stall during bulb screwing, predicting little or no motion.

\experimentparagraph{O8: Visual augmentation and broader controller training improve success}
\label{sec:stageO8}%
Without planner visual augmentation, final-stage SR is 0\%
on both tasks, although Bulb Screw reaches 15\%. Training the controller only
on the target task's retargeted trajectories, without synthetic motion augmentation,
reduces mean SR from 87.5\% to 52.5\%: Coin Lift falls from 85\% to 70\%,
and Bulb Light from 90\% to 35\%. The larger drop on bulb installation
suggests that this task benefits especially from broader controller training.
This comparison changes both motion augmentation and training-data coverage.

\section{Conclusion}
\label{sec:conclusion}

CIMI connects natural human demonstrations to robot hands through
contact-grounded observations, retargeting, and action representations.
Across six hand--task pairs on LEAP and Wuji, it improves mean success by
at least 50 percentage points over baseline policies. These results support
shared contact surfaces as a practical interface for transferring
contact-sensitive skills across different hand kinematics. We provide
the hardware designs and software to support broader robot-free demonstration
collection. Extending CIMI to bimanual and mobile manipulation offers a path
toward transferring a wider range of everyday skills.

\ifdefined\CIMIPublicVersion
  \newcommand{\BIBdecl}{\setlength{\baselineskip}{0.98\baselineskip}}
\fi
\begingroup

\endgroup

\setcounter{section}{0}
\setcounter{subsection}{0}
\setcounter{subsubsection}{0}
\setcounter{equation}{0}
\setcounter{figure}{0}
\setcounter{table}{0}
\renewcommand{\theequation}{S\arabic{equation}}
\renewcommand{\thefigure}{S\arabic{figure}}
\renewcommand{\thetable}{S\arabic{table}}
\renewcommand{\theHsection}{supp.\arabic{section}}
\renewcommand{\theHsubsection}{\theHsection.\arabic{subsection}}
\renewcommand{\theHsubsubsection}{\theHsubsection.\arabic{subsubsection}}
\renewcommand{\theHequation}{supp.\arabic{equation}}
\renewcommand{\theHfigure}{supp.\arabic{figure}}
\renewcommand{\theHtable}{supp.\arabic{table}}

\definecolor{cimiblueedge}{HTML}{3D82F6}
\definecolor{cimigreenedge}{HTML}{38A85B}
\definecolor{cimiorangeedge}{HTML}{F28C28}
\definecolor{cimiyellowedge}{HTML}{D6A900}
\setcounter{dbltopnumber}{1}
\renewcommand{\dbltopfraction}{0.65}
\renewcommand{\dblfloatpagefraction}{0.95}
\setcounter{tocdepth}{2}

\makeatletter
\let\maketitle\CIMIMakeTitle
\let\@maketitle\CIMIInternalMakeTitle
\renewcommand{\tableofcontents}{\section*{Supplementary Contents}\@starttoc{stoc}}
\renewcommand{\l@section}[2]{%
  \begingroup\bfseries\@dottedtocline{1}{0em}{1.6em}{#1}{#2}\endgroup}
\renewcommand{\l@subsection}{\@dottedtocline{2}{1.6em}{2.8em}}
\let\CIMIAddContentsLine\addcontentsline
\renewcommand{\addcontentsline}[3]{%
  \def\CIMIContentsType{#1}\def\CIMIToCType{toc}%
  \ifx\CIMIContentsType\CIMIToCType
    \CIMIAddContentsLine{stoc}{#2}{#3}%
  \else
    \CIMIAddContentsLine{#1}{#2}{#3}%
  \fi}
\makeatother

\title{\LARGE \bf
Contact Imitation Interface:\\
Transferring Dexterous Skills by Showing How Contact Unfolds\\[0.4ex]
\large Supplementary Material
}
\maketitle
\phantomsection
\makeatletter
\def\@currentlabel{S}
\def\@currentlabelname{Supplementary Material}
\makeatother
\label{sec:supplement}
\thispagestyle{plain}
\pagestyle{plain}
\tableofcontents

\section{Demonstration Interface}
This section expands the capture hardware (D1), visual tracking (D2),
contact-surface retargeting (D3), and context-camera mounting used for
observation conversion (D4) in the main paper.

\subsection{Contact-Grounded Capture Hardware}
\label{sec:supp_hardware}
This section describes the hardware design and fabrication of the demonstration
interface in D1 of the main paper. The design combines matched fingertip modules,
wider-view stereo cameras, and dedicated tracking views while keeping the custom
mechanical components straightforward to reproduce.

\begin{figure*}[t]
\centering
\ifdefined\CIMIInlineFigureSource
\begin{tikzpicture}[
  baseframe/.style={rounded corners=2pt,align=center,minimum width=34mm,
    minimum height=10mm,inner sep=3pt,font=\small,line width=0.9pt,
    drop shadow={opacity=0.13,shadow xshift=0.7pt,shadow yshift=-0.7pt}},
  camera/.style={baseframe,draw=cimiblueedge,fill=cimiblue},
  target/.style={baseframe,draw=cimiorangeedge,fill=cimiorange},
  surface/.style={baseframe,draw=cimigreenedge,fill=cimigreen},
  estimated/.style={-{Latex[length=2.2mm,width=1.5mm]},draw=cimiyellowedge,
    densely dashed,line width=1.35pt},
  calibrated/.style={-{Latex[length=2.2mm,width=1.5mm]},draw=cimiorangeedge,
    line width=1.35pt}]

\node[font=\small\bfseries,text=black!70] at (-4.8,0.65) {Transform type};
\draw[estimated] (-3.25,0.65) -- (-1.85,0.65);
\node[anchor=west,font=\small,text=black!75] at (-1.65,0.65) {estimated from video};
\draw[calibrated] (1.70,0.65) -- (3.10,0.65);
\node[anchor=west,font=\small,text=black!75] at (3.30,0.65) {fixed calibration};

\node[camera] (e) at (0,-0.55) {External tracking camera\\$E$};
\node[target] (d) at (0,-1.85) {Dorsal fiducial\\$D_H$};
\node[camera] (cw) at (-3.25,-3.15) {Thumb-web tracking camera\\$C_{H,\mathrm{web}}$};
\node[camera] (cm) at (3.95,-3.15) {Middle-finger tracking camera\\$C_{H,\mathrm{middle}}$};
\node[target] (mt) at (-5.25,-4.45) {Thumb fiducial\\$M_{H,\mathrm{thumb}}$};
\node[target] (mi) at (-1.25,-4.45) {Index fiducial\\$M_{H,\mathrm{index}}$};
\node[target] (mm) at (3.95,-4.45) {Middle-finger fiducial\\$M_{H,\mathrm{middle}}$};
\node[surface] (st) at (-5.25,-5.75) {Thumb contact surface\\$S_{H,\mathrm{thumb}}$};
\node[surface] (si) at (-1.25,-5.75) {Index contact surface\\$S_{H,\mathrm{index}}$};
\node[surface] (sm) at (3.95,-5.75) {Middle-finger contact surface\\$S_{H,\mathrm{middle}}$};

\draw[estimated] (e) -- (d);
\draw[calibrated] (d) -- (cw);
\draw[calibrated] (d) -- (cm);
\draw[estimated] (cw) -- (mt);
\draw[estimated] (cw) -- (mi);
\draw[estimated] (cm) -- (mm);
\draw[calibrated] (mt) -- (st);
\draw[calibrated] (mi) -- (si);
\draw[calibrated] (mm) -- (sm);
\end{tikzpicture}
\else
\includegraphics{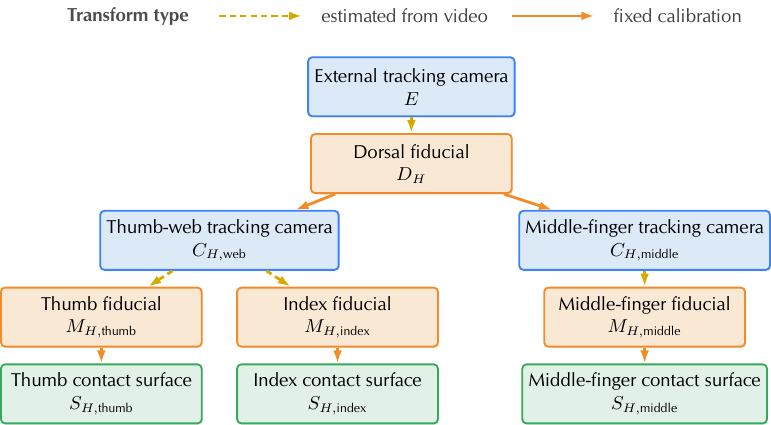}
\fi
\caption{\textbf{Transformation tree for visual contact-surface tracking.} Each edge
specifies the child frame's pose in its parent frame. Dashed yellow edges are
estimated from videos; solid orange edges are fixed mounting calibrations.}
\label{fig:tracking_tree}
\end{figure*}

\subsubsection{Why Visual Tracking?}
\noindent\textbf{Direct contact-surface calibration.}
Visual tracking directly measures our rigid contact modules using known target
geometry and calibrated target-to-surface transforms. A glove-plus-tracker
system instead requires additional registration from its calibrated hand model
to these contact surfaces~\cite{manusMetaglovesSpecs,manusHandMeasurement}.

\par\smallskip
\noindent\textbf{Contact-camera compatibility.}
In our tests with MANUS Metagloves Pro, MANUS Core~3, and SDK access,
simultaneous contact-camera operation could degrade image quality or shift the
estimated hand pose, consistent with electromagnetic interference; MANUS also
warns of this sensitivity~\cite{manusHandMeasurement}.

\par\smallskip
\noindent\textbf{Lower hardware cost.}
The vendor lists Metagloves Pro at
EUR~4,500~\cite{manusProductPricing}, while our capture-interface hardware totals
\$778.95 (Table~\ref{tab:supp_bom}). The listed glove price does not establish
the cost of our tested software/SDK configuration.

\subsubsection{Hardware Design Details}
\noindent\textbf{Matched fingertip modules.}
Each human or robot fingertip module uses one HBVCAM-2307-FPC82 V11 USB camera.
The camera measures approximately
$8.6\times8.6\times5.4$\,mm and has a $120^\circ$ field of view. It is mounted
at the fingertip root and oriented tangentially toward the local contact region.
The human and robot variants preserve the same nominal contact-surface geometry
and camera placement, with hand-specific 3D-printed mounting structures.
A silicone sleeve provides compliant contact without obstructing the camera
view. We apply one layer of friction tape on the human contact surface and two
on the robot surface; the additional robot-side thickness supports contact
compression, as discussed in Sec.~\ref{sec:supp_pose_force}.

\par\smallskip
\noindent\textbf{Context and tracking cameras.}
Two Orbbec Gemini 305 stereo cameras are mounted at the wrist and near the
thumb--index web space. Together they provide the four context-image streams
used for reconstruction and viewpoint conversion. One DCXG500 RGB camera
tracks the middle-finger fiducial, and an external DCXG500 camera tracks the
dorsal fiducial on the back of the hand. The fingertip targets, dorsal target,
and tracking-camera mounts are rigidly attached so that their fixed
transformations can be calibrated once and reused across recordings.

\par\smallskip
\noindent\textbf{Fabrication and assembly.}
We first 3D print the fingertip bodies, camera holders, hand-specific adapters,
tracking-camera mounts, and fiducial supports. We then install one contact camera
and one silicone sleeve on each fingertip module, attach the AprilCube targets,
and mount the modules to the human interface or robot hand. Finally, we install
the two Gemini 305 cameras and the auxiliary tracking cameras, calibrate camera
intrinsics and the fixed transformations in Fig.~\ref{fig:tracking_tree}, and
verify all nine streams at the 10\,Hz recording rate. The printed geometry fixes
the nominal camera--contact-surface placement, which is shared between human
demonstration and robot execution.

\subsubsection{Bill of Materials}
Table~\ref{tab:supp_bom} lists the principal components for one three-finger
demonstration interface in USD; the contact-camera price follows the FingerEye
BOM. Printed parts, silicone sleeves, friction tape, and printed fiducials have a combined
estimated material cost of \$7.00. Costs exclude labor, printer depreciation,
cables, and the host computer.
\begin{table}[ht]
\centering
\footnotesize
\setlength{\tabcolsep}{3pt}
\begin{tabularx}{\columnwidth}{@{}Ycr@{}}
\toprule
Component & Qty. & Cost (USD) \\
\midrule
HBVCAM-2307-FPC82 V11 contact camera & 3 & \$28.25 each \\
Orbbec Gemini 305 stereo camera & 2 & \$259.18 each \\
DCXG500 RGB tracking camera (middle-finger and external views) & 2 & \$84.42 each \\
Fabrication materials (printed parts, silicone sleeves, friction tape, and fiducials) & 1 set & \$7.00 (est.) \\
\midrule
\multicolumn{2}{@{}l}{Estimated hardware total} & \$778.95 \\
\bottomrule
\end{tabularx}
\par\vspace{4pt}
\caption{Bill of materials for one three-finger demonstration interface. The
total covers three contact cameras, two Gemini 305 cameras, two auxiliary
tracking cameras, and the combined fabrication-material estimate.}
\label{tab:supp_bom}
\end{table}

\subsection{Visual Contact-Surface Tracking}
\label{sec:supp_tracking}
This section details the visual estimator and calibrated transformation tree
used to extract contact-surface trajectories in D2 of the main paper.

\begin{figure*}[t]
\centering
\includegraphics[width=\textwidth]{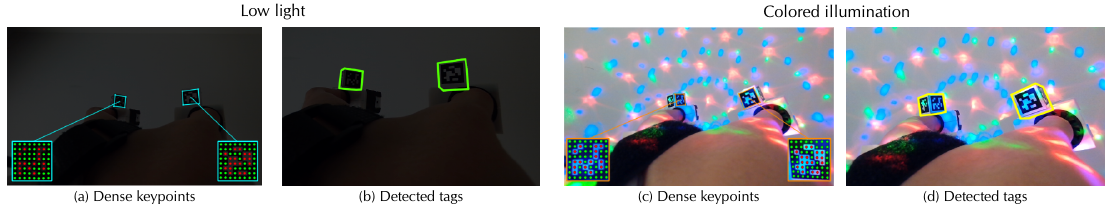}
\caption{\textbf{Qualitative lighting robustness of fiducial detection.}
(a,b) Low-light examples. (c,d) Examples with spatially varying colored
illumination. For each condition, the left panel shows dense keypoints with
enlarged tag-grid insets, and the right panel shows detected tag outlines.
All detection overlays are preserved from the original images.}
\label{fig:supp_lighting_robustness}
\end{figure*}

\subsubsection{Fiducial Targets and Tracking Frames}
We use AprilTag markers \cite{olson2011apriltag} on AprilCube targets
\cite{park2026aprilcube} and detect their dense grid keypoints with DeepTag
\cite{zhang2023deeptag}. Each fingertip target is rigidly attached to its
contact module. A dorsal fiducial on the back of the hand provides a moving
reference that connects the hand-mounted tracking cameras to the external
tracking-camera frame.

Let ${}^{A}\mathbf T^{B}_t\in\mathrm{SE}(3)$ map coordinates from frame $B$
to frame $A$ at time $t$. The subscript $H$ denotes the human demonstration
interface. We use $E$ for the fixed external tracking-camera frame, $D_H$ for
the moving dorsal fiducial frame, $M_{H,i}$ for fingertip $i$'s target frame,
and $S_{H,i}$ for its contact-surface frame. The index $j(i)$ identifies the
hand-mounted tracking camera that observes fingertip $i$, with frame
$C_{H,j(i)}$.
The thumb and index targets share a thumb-web tracking view, while the middle
finger uses a separate tracking view:
\[
\begin{aligned}
j(\mathrm{thumb})=j(\mathrm{index})&=\mathrm{web},\\
j(\mathrm{middle})&=\mathrm{middle}.
\end{aligned}
\]
These tracking views estimate target poses; the contact cameras described in
D1 provide the policy's contact observations.

\subsubsection{Full Contact-Surface Transformation Tree}
Figure~\ref{fig:tracking_tree} shows the complete tree for the three-finger
capture configuration. Each contact-surface pose in the external camera frame
is obtained by composing two tracked poses and two fixed mounting transforms:
\begin{equation}
{}^{E}\mathbf T^{S_{H,i}}_t
={}^{E}\mathbf T^{D_H}_t
{}^{D_H}\mathbf T^{C_{H,j(i)}}
{}^{C_{H,j(i)}}\mathbf T^{M_{H,i}}_t
{}^{M_{H,i}}\mathbf T^{S_{H,i}}.
\label{eq:surface_tracking}
\end{equation}
The external view estimates ${}^{E}\mathbf T^{D_H}_t$, and the hand-mounted
view estimates ${}^{C_{H,j(i)}}\mathbf T^{M_{H,i}}_t$.
The calibrated transforms ${}^{D_H}\mathbf T^{C_{H,j(i)}}$ and
${}^{M_{H,i}}\mathbf T^{S_{H,i}}$ describe the camera mounts and the
contact module's geometry, respectively, and remain fixed during recording.
Calibrated camera intrinsics and target keypoint coordinates specified in
physical units make the recovered translations metric.

The four local corners of each contact surface are known from its geometry.
For corner $\ell$, let $\bar{\mathbf v}_{i,\ell}=(v_x,v_y,v_z,1)^\top$ be
its homogeneous coordinates in $S_{H,i}$. The demonstration target used for
contact-surface retargeting in D3 is
\begin{equation}
\mathbf p^H_{i,\ell,t}
=\left[{}^{E}\mathbf T^{S_{H,i}}_t\bar{\mathbf v}_{i,\ell}\right]_{1:3},
\end{equation}
where $[\cdot]_{1:3}$ retains the three Cartesian coordinates.

\subsubsection{Joint Multi-Tag Pose Estimation}
DeepTag provides dense image keypoints on each detected tag. We map their
canonical grid locations to 3D points using the tag's known dimensions,
decoded orientation, and placement on the rigid target. We aggregate these
3D--2D correspondences across visible tags and solve for one target pose
with multi-tag Perspective-$n$-Point (PnP) \cite{xu2026fingereye}.

For a target frame $M$ observed by camera $C$, let $\mathbf x_k$ be a known
3D keypoint in $M$ and $\mathbf u_k$ its detected image location. The pose
refinement minimizes the joint reprojection error over accepted keypoints
$\mathcal I$:
\begin{equation}
{}^{C}\mathbf T^{M}_t
=\operatorname*{argmin}_{\mathbf T\in\mathrm{SE}(3)}
\sum_{k\in\mathcal I}
\left\|\pi_{\mathbf K,\mathbf d}
\left([\mathbf T\bar{\mathbf x}_k]_{1:3}\right)-\mathbf u_k\right\|_2^2,
\end{equation}
where $\bar{\mathbf x}_k=(\mathbf x_k^\top,1)^\top$, and
$\pi_{\mathbf K,\mathbf d}$ projects a camera-frame point using calibrated
intrinsics $\mathbf K$ and distortion parameters $\mathbf d$.

For observations spanning multiple target faces, the implementation initializes
the pose with SQPnP inside RANSAC, rejects inconsistent keypoints and tags,
and refines the pose with Levenberg--Marquardt using the retained correspondences.
When only one face is visible, we use planar pose estimation with reprojection
and face-orientation checks. Joint estimation lets visible tags contribute
redundant constraints when other tags are occluded, provided enough valid
keypoints remain. The same estimator supplies both tracked transforms in
Eq.~\eqref{eq:surface_tracking}.

\subsubsection{Detection under Challenging Illumination}
Figure~\ref{fig:supp_lighting_robustness} shows qualitative examples of
fiducial detection under low light and spatially varying colored illumination.
The dense keypoint overlays and detected tag outlines illustrate successful
detections under both lighting conditions.

\begin{figure*}[t]
\centering
\ifdefined\CIMIInlineFigureSource
\begin{tikzpicture}[
  x=1cm,y=1cm,
  block/.style={rounded corners=1.5pt,align=center,minimum height=7.5mm,
    inner sep=2pt,font=\fontsize{6.7}{7.8}\selectfont,line width=0.6pt},
  view/.style={block,draw=cimiblueedge,fill=cimiblue,text width=24mm},
  optionalview/.style={view,densely dashed},
  vision/.style={block,draw=cimiblueedge,fill=cimiblue},
  geometry/.style={block,draw=cimigreenedge,fill=cimigreen},
  flow/.style={block,draw=cimiorangeedge,fill=cimiorange},
  output/.style={block,draw=cimigreenedge,fill=cimigreen},
  arrow/.style={-{Latex[length=1.45mm,width=1mm]},draw=black!60,
    line width=0.6pt,shorten >=0.6pt,shorten <=0.6pt},
  optionalarrow/.style={arrow,densely dashed},
  legend/.style={anchor=west,font=\fontsize{6.7}{7.8}\selectfont}]

\node[draw=cimiblueedge,fill=cimiblue,minimum size=2mm,inner sep=0pt]
  at (-8.10,2.95) {};
\node[legend] at (-7.90,2.95) {RGB processing};
\node[draw=cimigreenedge,fill=cimigreen,minimum size=2mm,inner sep=0pt]
  at (-3.85,2.95) {};
\node[legend] at (-3.65,2.95) {Geometry and conditioning};
\node[draw=cimiorangeedge,fill=cimiorange,minimum size=2mm,inner sep=0pt]
  at (2.55,2.95) {};
\node[legend] at (2.75,2.95) {Flow prediction};

\node[optionalview] (contact) at (-7.0,2.00)
  {Contact (optional)\\$0$ or $F$ RGB views\\$240\times320$};
\node[view] (wrist) at (-7.0,1.05)
  {Wrist context RGB\\$1$ view; $320\times512$};
\node[optionalview] (web) at (-7.0,0.10)
  {Web RGB (optional)\\$0$ or $1$ view\\$320\times512$};
\node[vision,text width=26mm] (radio) at (-3.75,1.05)
  {Frozen C-RADIOv3-B\\context resize: $240\times384$\\$2304$-D per view};
\node[vision,text width=22mm] (project) at (-0.65,1.05)
  {$V$ view-specific projections\\$2304\rightarrow768$};
\node[vision,text width=27mm] (fuse) at (2.45,1.05)
  {Concatenate + fuse\\$768V\rightarrow768\rightarrow768\rightarrow768$\\GELU after first two layers};
\node[geometry,text width=25mm] (condition) at (5.90,1.05)
  {Planner condition $\mathbf o_t$\\$d_c=768+9(F-1)$};
\node[geometry,text width=25mm] (prop) at (5.90,-0.10)
  {Relative contact poses\\$\mathbf s_t^{\mathrm{prop}}\in\mathbb R^{9(F-1)}$};

\draw[optionalarrow] (contact.east) -- ([yshift=2.5mm]radio.west);
\draw[arrow] (wrist.east) -- (radio.west);
\draw[optionalarrow] (web.east) -- ([yshift=-2.5mm]radio.west);
\draw[arrow] (radio.east) -- (project.west);
\draw[arrow] (project.east) -- (fuse.west);
\draw[arrow] (fuse.east) -- (condition.west);
\draw[arrow] (prop.north) -- (condition.south);

\node[flow,text width=24mm] (state) at (-7.0,-1.15)
  {Flow state\\$\mathrm{vec}(\mathbf Z_{t,s})\in\mathbb R^{9FH}$};
\node[flow,text width=24mm] (time) at (-7.0,-2.10)
  {Flow time $s$\\sinusoidal $128$\\$128\rightarrow256\rightarrow128$};
\node[flow,text width=24mm] (condproj) at (-7.0,-3.05)
  {Condition $\mathbf o_t$: $d_c\rightarrow768$\\LayerNorm + GELU};
\node[flow,text width=25mm] (inputproj) at (-3.75,-2.10)
  {Concatenate + project\\$9FH+896\rightarrow768$};
\node[flow,text width=27mm] (blocks) at (-0.55,-2.10)
  {$6\times$ time-residual blocks\\LN; time $128\rightarrow768$\\$768\rightarrow3072\rightarrow768$};
\node[flow,text width=24mm] (velocity) at (2.65,-2.10)
  {LN + linear + reshape\\$768\rightarrow9FH$\\$v_\theta\in\mathbb R^{H\times9F}$};
\node[flow,text width=17mm] (integrate) at (5.35,-2.10)
  {Euler\\integration\\$3$ flow steps};
\node[output,text width=21mm] (prediction) at (7.85,-2.10)
  {Predicted trajectory\\$\widehat{\mathbf U}_t\in\mathbb R^{H\times9F}$};

\draw[arrow] (state.east) -- ([yshift=2.5mm]inputproj.west);
\draw[arrow] (time.east) -- (inputproj.west);
\draw[arrow] (condproj.east) -- ([yshift=-2.5mm]inputproj.west);
\draw[arrow] (inputproj.east) -- (blocks.west);
\draw[arrow] (blocks.east) -- (velocity.west);
\draw[arrow] (velocity.east) -- (integrate.west);
\draw[arrow] (integrate.east) -- (prediction.west);

\node[anchor=west,font=\fontsize{6.7}{7.8}\selectfont\itshape,text=black!65]
  at (-8.25,-3.75)
  {Only the camera views selected for each policy are encoded. Both paths use the same condition $\mathbf o_t$.};
\end{tikzpicture}
\else
\includegraphics{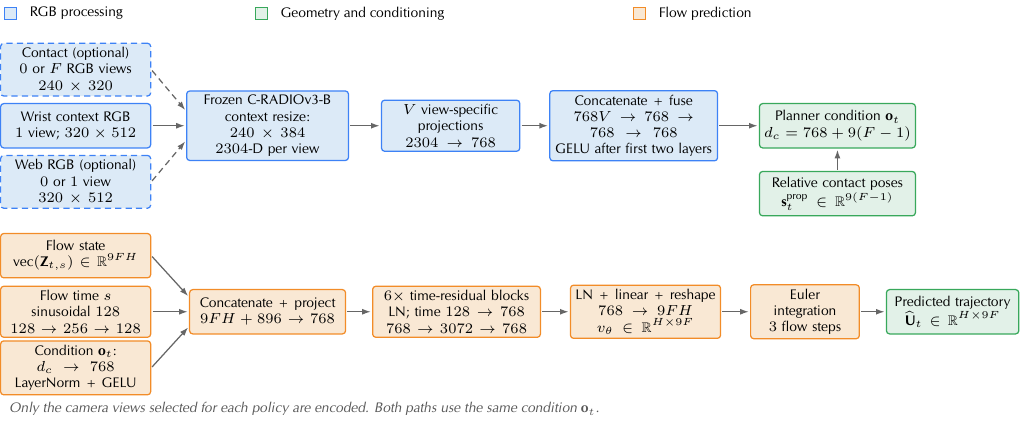}
\fi
\caption{\textbf{Contact-surface planner architecture.} The upper path encodes
the $V$ configured RGB views and the current inter-finger geometry into a
condition vector. The lower path evaluates the conditional velocity network and
integrates it for three Euler steps to predict an $H$-step contact-surface
trajectory. Blue groups RGB processing, green groups geometric inputs,
conditioning, and the predicted trajectory, and orange groups flow prediction.
Dashed boxes and arrows identify optional camera inputs.
Here $F$ is the number of controlled fingers and each contact-surface pose has
nine coordinates (three translation and six rotation coordinates).}
\label{fig:supp_planner_architecture}
\end{figure*}

\subsection{Contact-Surface Retargeting}
\label{sec:supp_retargeting}
The retargeting implementation optimizes the robot root pose and active
finger joints over each recorded sequence. Camera alignment initializes the
root pose; it is not an additional objective in the free-wrist variant.
Inactive joints remain fixed. The objective combines contact-corner error with
first-order temporal regularization.

\subsubsection{Coordinates and Objective}
The numerical optimization expresses both human targets and robot predictions
in the moving dorsal fiducial frame $D_H$, which is rigidly attached to the
back of the hand. The estimated transform ${}^{E}\mathbf T^{D_H}_t$ maps this
moving frame into the fixed external tracking-camera frame $E$; applying the
same transform to both targets and predictions leaves their instantaneous
Euclidean contact error unchanged. Let
$\widetilde{\mathbf X}_t=(\widetilde{\mathbf R}_t,\widetilde{\mathbf x}_t)$
be the robot root pose in this frame, and let $\mathcal V$ contain all valid
contact-corner tuples $(t,i,\ell)$. All retained active corners are weighted
equally; inactive or invalid corners are omitted. The contact term is
\begin{equation}
\mathcal L_{\mathrm{contact}}=
\frac{1}{|\mathcal V|}\sum_{(t,i,\ell)\in\mathcal V}
\left\|\widetilde{\mathbf p}^{R}_{i,\ell,t}
-\widetilde{\mathbf p}^{H}_{i,\ell,t}\right\|_2^2.
\end{equation}
Positions are expressed in meters. For consecutive retained frames, define
$\delta\mathbf q_t=\mathbf q_t-\mathbf q_{t-1}$,
$\delta\mathbf x_t=\widetilde{\mathbf x}_t-\widetilde{\mathbf x}_{t-1}$,
and
$\delta\boldsymbol\omega_t=
\log(\widetilde{\mathbf R}_{t-1}^{\top}\widetilde{\mathbf R}_t)$,
where the rotation logarithm returns a three-dimensional rotation vector.
With temporal edge weights $e_t$ and
$Z_e=\max(\sum_t e_t,1)$, the temporal term is
\begin{equation}
\begin{aligned}
\mathcal R=\frac{1}{Z_e}\sum_t e_t\Bigl[
&\lambda_q\|\delta\mathbf q_t\|_2^2
+\lambda_x\|\delta\mathbf x_t\|_2^2\\
&+\lambda_R\|\delta\boldsymbol\omega_t\|_2^2\Bigr].
\end{aligned}
\end{equation}
Unit conversions and per-dimension normalizations are absorbed into the
temporal coefficients.
We minimize $\mathcal L_{\mathrm{contact}}+\mathcal R$ subject to joint limits.
Temporal differences of the root pose are measured in the moving dorsal
fiducial frame $D_H$; the final robot trajectory is transformed to the fixed
external tracking-camera frame $E$.

\subsubsection{Optimizer Configuration}
Table~\ref{tab:supp_retarget_optimizer} summarizes the task-independent
configuration. The solver differentiates through robot forward kinematics,
while contact-module geometry remains fixed. Separate recordings have no
cross-recording temporal edges and are normalized independently.
\begin{table}[ht]
\centering
\footnotesize
\caption{Task-independent retargeting optimizer configuration.}
\label{tab:supp_retarget_optimizer}
\setlength{\tabcolsep}{4pt}
\begin{tabularx}{\columnwidth}{@{}p{0.34\columnwidth}Y@{}}
\toprule
Setting & Value \\
\midrule
Solver schedule & 4,000 projected Adam steps, then up to 1,000 L-BFGS-B steps \\
Optimization variables & Scaled to $[-1,1]$ \\
Floating-root bound & $\pm 0.20$\,m per axis from initialization \\
Temporal-gap handling & Edge weight $1/g$ for a $g$-frame interval; no edge across more than 16 missing frames \\
\bottomrule
\end{tabularx}
\end{table}

\subsection{Context-Camera Mount Optimization}
\label{sec:supp_camera_mounts}
For each robot hand, we fit one fixed mounting pose per context camera across
the collected demonstrations, holding the retargeted wrist and finger motion
fixed. Let $\mathbf B_t$ be the robot-palm pose and $\mathbf H_{j,t}$ the human
pose of camera $j$, expressed in a common reference frame after handedness
conversion. The mount $\mathbf M_j\in\mathrm{SE}(3)$ maps the left camera's
optical frame into the robot palm frame:
\begin{equation}
\mathbf M_j^*=\operatorname*{argmin}_{\mathbf M_j}
\sum_t w_t\,d_j(\mathbf B_t\mathbf M_j,\mathbf H_{j,t}).
\label{eq:supp_camera_mount_fit}
\end{equation}
Here $d_j$ penalizes displacement between the predicted and demonstrated camera
centers and disagreement between their orientations. The weights $w_t$ give
equal total weight to each task, divided equally among its recordings and then
their valid frames. The cameras are fitted independently; the LEAP wrist
camera additionally constrains its optical $+X$ axis to lie in the palm's
$XY$ plane. D4 uses these fitted mounts for viewpoint reprojection; physical
mounting calibration is performed separately. Contact-camera placement remains
fixed by the shared fingertip modules.

\subsection{Processing Efficiency and Data Quality}
\label{sec:supp_processing_efficiency}
Table~\ref{tab:supp_processing_time} covers videos to policy-ready features:
two contact views and nominal/augmented wrist views, using the augmentation
in Section~\ref{sec:supp_planner}.

\begin{table}[ht]
\centering
\footnotesize
\caption{Coin-picking processing on an RTX 4090 D (10 demonstrations,
557 frames). Serial stage wall times include initialization and output writing.}
\label{tab:supp_processing_time}
\setlength{\tabcolsep}{4pt}
\begin{tabularx}{\columnwidth}{@{}Xr@{}}
\toprule
Stage & Time (s) \\
\midrule
Visual tracking (image2cube) & 57 \\
Contact-surface retargeting (cube2qpos) & 43 \\
Viewpoint reprojection and robot overlay & 72 \\
Image augmentation and RADIO caching & 52 \\
\midrule
Total & 224 \\
\bottomrule
\end{tabularx}
\end{table}

Table~\ref{tab:supp_data_retention} reports retention among stored task
demonstrations, not all collection attempts. Coin rejects one recording shorter
than 16 steps; bulb rejects two below 50\% valid frames. Invalid frames remain
masked during learning.

\begin{table}[ht]
\centering
\footnotesize
\caption{Data retention for audited LEAP task datasets. Valid-frame coverage
is weighted by recording length, not averaged over recordings.}
\label{tab:supp_data_retention}
\setlength{\tabcolsep}{4pt}
\begin{tabular}{@{}lrrr@{}}
\toprule
Task & Retained / stored & Retention & Valid frames \\
\midrule
Coin & 209 / 210 & 99.5\% & 99.3\% \\
Letter & 205 / 205 & 100.0\% & 95.5\% \\
Bulb & 115 / 117 & 98.3\% & 79.0\% \\
\bottomrule
\end{tabular}
\end{table}

\section{Learning Interface}
We first detail the contact-surface planner (L2), then the controller (L3),
followed by execution, reuse across tasks, policy-camera preprocessing, and
computational cost.

\subsection{Contact-Surface Planner}
\label{sec:supp_planner}

\subsubsection{Network Architecture and Dimensions}
Figure~\ref{fig:supp_planner_architecture} shows the planner's visual conditioning
and flow-prediction paths with their dimensions;
Table~\ref{tab:supp_planner_settings} lists normalization and training settings.
Here $F$ is the number of controlled fingers,
$V$ the number of RGB views, and $H$ the prediction horizon.

The configured view set defines $V$: \textbf{\texttt{CTX}} uses the task's context views, while
\textbf{\texttt{CTX+CC}} additionally includes one contact view per controlled finger. The wrist
context view is used for all reported tasks, whereas the web-space context view
is enabled only for bulb installation. Separate policies are trained for each
view set. Only the selected views are encoded and concatenated;
unused cameras are not inputs to that policy.

\begin{figure*}[t]
\centering
\includegraphics[width=\textwidth]{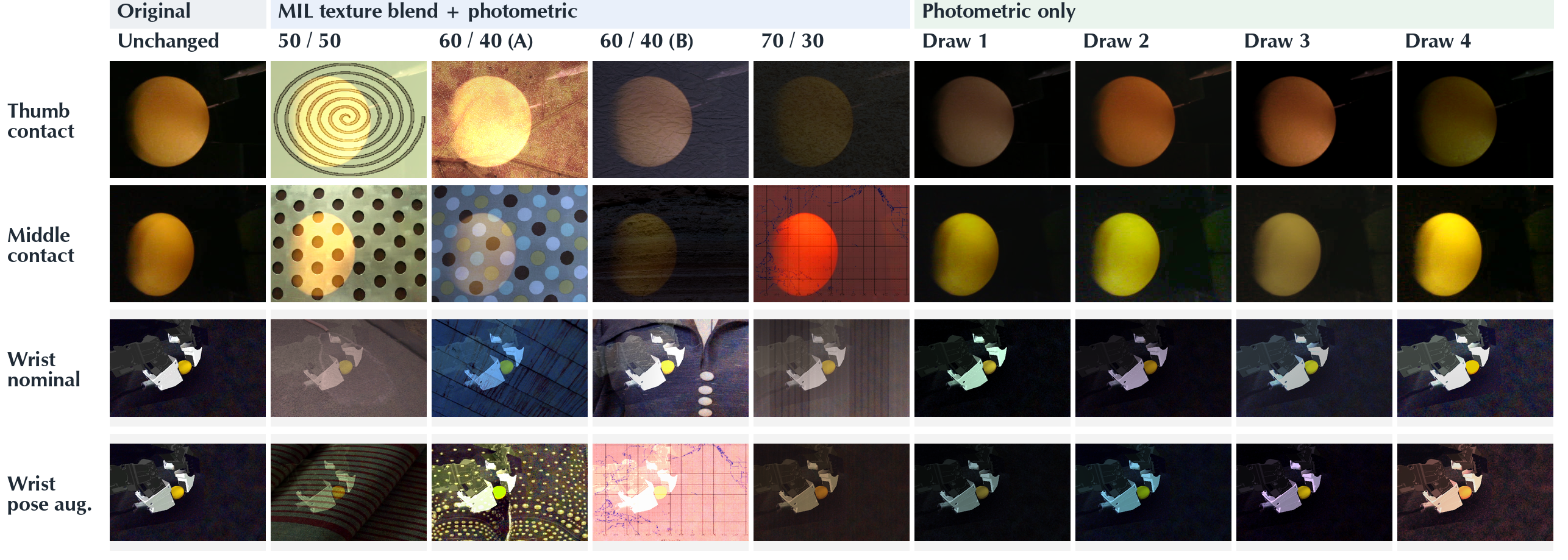}
\caption{\textbf{Visual augmentation for coin picking.} Rows show the thumb
and middle-finger contact views, followed by wrist context views with nominal
and perturbed camera extrinsics. The perturbed wrist render also varies robot
lighting and materials. Each base image has four whole-image MIL blends with
photometric augmentation and four photometric-only variants. Blend columns
give original/texture percentages; the two 60/40 candidates use distinct textures.}
\label{fig:supp_visual_augmentation}
\end{figure*}

\subsubsection{Flow Matching and Conditioning}
Given the combined observation
$\mathbf o_t=(\mathbf z_t^{\mathrm{vis}},\mathbf s_t^{\mathrm{prop}})$, the
conditional planner $\pi_\theta(\mathbf U_t\mid\mathbf o_t)$ is implemented with
a velocity field $v_\theta$ that transforms a source sample
$\boldsymbol\epsilon\sim\mathcal U([-1,1]^{H\times9F})$ into a contact-motion
chunk. For flow time $s\in[0,1]$, generation follows
\begin{equation}
\begin{aligned}
\frac{d\mathbf Z_{t,s}}{ds}
&=v_\theta(\mathbf Z_{t,s},s\mid\mathbf o_t),\\
\mathbf Z_{t,0}&=\boldsymbol\epsilon,\qquad
\widehat{\mathbf U}_t=\mathbf Z_{t,1}.
\end{aligned}
\label{eq:supp_planner_integration}
\end{equation}
Here $\mathbf Z_{t,s}$ is the evolving chunk, and
$\widehat{\mathbf U}_t=[\widehat{\mathbf u}_{t,1},\ldots,
\widehat{\mathbf u}_{t,H}]$ is the predicted relative contact-surface trajectory. We train
$v_\theta$ with conditional flow matching \cite{lipman2023flowmatching}.
For demonstration chunk
$\mathbf U_t=[\mathbf u_{t,1},\ldots,\mathbf u_{t,H}]$, let
$\mathbf U_{t,s}=(1-s)\boldsymbol\epsilon+s\mathbf U_t$.
Training minimizes
\begin{equation}
\mathcal L_{\mathrm{FM}}=\mathbb E\left[
\|v_\theta(\mathbf U_{t,s},s\mid\mathbf o_t)
-(\mathbf U_t-\boldsymbol\epsilon)\|_2^2\right].
\label{eq:flow_matching}
\end{equation}

The planner's proprioceptive input consists of non-thumb contact-surface poses
relative to the thumb. These poses are computed by forward kinematics from
retargeted joints during training and measured robot joints at execution.
Only valid future steps contribute to the planner's flow-matching loss;
padded horizon entries are masked out. The controller is trained separately,
and its parameters remain fixed during planner training.

\subsubsection{Optimization Settings}
LEAP and Wuji share the planner settings in Table~\ref{tab:supp_planner_settings}.

\begin{table}[t]
\centering
\footnotesize
\caption{Planner normalization and training settings.}
\label{tab:supp_planner_settings}
\setlength{\tabcolsep}{3pt}
\renewcommand{\arraystretch}{1.08}
\begin{tabularx}{\columnwidth}{@{}P{0.35\columnwidth}Y@{}}
\toprule
Setting & Value \\
\midrule
Contact-surface scaling & Translation $\times20$; six rotation coordinates unchanged \\
Direct baseline scaling & Global 2nd--98th percentile for non-rotation coordinates; rotation unchanged \\
Dropout & 0 \\
Batch and budget & 256; up to 63,000 optimizer updates \\
AdamW & LR $10^{-4}$; $\beta=(0.95,0.999)$; $\epsilon=10^{-8}$; weight decay $10^{-6}$ \\
LR and EMA schedules & 500 warm-up updates then cosine decay; EMA inverse-gamma 1, power 0.75, maximum 0.9999 \\
\bottomrule
\end{tabularx}
\end{table}

\begin{figure*}[t]
\centering
\ifdefined\CIMIInlineFigureSource
\begin{tikzpicture}[
  x=1cm,y=1cm,
  block/.style={rounded corners=1.5pt,align=center,minimum height=7.5mm,
    inner sep=2pt,font=\fontsize{6.7}{7.8}\selectfont,line width=0.6pt},
  input/.style={block,draw=cimigreenedge,fill=cimigreen},
  geometry/.style={block,draw=cimiblueedge,fill=cimiblue},
  network/.style={block,draw=cimiorangeedge,fill=cimiorange},
  output/.style={block,draw=cimigreenedge,fill=cimigreen},
  arrow/.style={-{Latex[length=1.45mm,width=1mm]},draw=black!60,
    line width=0.6pt,shorten >=0.6pt,shorten <=0.6pt},
  legend/.style={anchor=west,font=\fontsize{6.7}{7.8}\selectfont}]

\node[draw=cimigreenedge,fill=cimigreen,minimum size=2mm,inner sep=0pt]
  at (-8.15,1.70) {};
\node[legend] at (-7.95,1.70) {Inputs and outputs};
\node[draw=cimiblueedge,fill=cimiblue,minimum size=2mm,inner sep=0pt]
  at (-3.10,1.70) {};
\node[legend] at (-2.90,1.70) {Geometric feature construction};
\node[draw=cimiorangeedge,fill=cimiorange,minimum size=2mm,inner sep=0pt]
  at (3.55,1.70) {};
\node[legend] at (3.75,1.70) {Learned decoder};

\node[input,text width=19mm] (plan) at (-7.45,0.60)
  {Relative\\surface plan\\$\widehat{\mathbf U}_t\in\mathbb R^{H\times9F}$};
\node[input,text width=19mm] (qpos) at (-7.45,-0.60)
  {Current joints\\$\mathbf q_t\in\mathbb R^{4F}$};
\node[geometry,text width=22mm] (corners) at (-4.95,0)
  {FK + relative poses\\Four corners per surface\\$\widehat{\mathbf P}_t\in\mathbb R^{H\times12F}$};
\node[geometry,text width=27mm] (features) at (-2.05,0)
  {Geometric features\\frames: $12F$\\relative frames: $12(F-1)$\\joints: $4F$; horizon: $1$\\$d_x=28F-11$\\$45$ ($F=2$); $73$ ($F=3$)};
\node[network,text width=16mm] (project) at (0.60,0)
  {Input\\projection\\$d_x\rightarrow768$\\LN + GELU};
\node[network,text width=23mm] (blocks) at (3.00,0)
  {$8\times$ residual blocks\\LN + GELU\\$768\rightarrow1536\rightarrow768$\\skip connections};
\node[network,text width=17mm] (head) at (5.35,0)
  {Output head\\LN + linear\\$768\rightarrow9+4F$\\zero-initialized};
\node[output,text width=22mm] (commands) at (7.70,0)
  {Commands\\$H\times(9+4F)$\\wrist scale: $0.15$\,m\\rotation: $6$\\joints clipped to limits};

\draw[arrow] (plan.east) -- ([yshift=2.5mm]corners.west);
\draw[arrow] (qpos.east) -- ([yshift=-2.5mm]corners.west);
\draw[arrow] (corners.east) -- (features.west);
\draw[arrow] (features.east) -- (project.west);
\draw[arrow] (project.east) -- (blocks.west);
\draw[arrow] (blocks.east) -- (head.west);
\draw[arrow] (head.east) -- (commands.west);
\draw[arrow,rounded corners=2pt] (qpos.south) -- (-7.45,-1.35)
  -- (-2.05,-1.35) -- (features.south);

\node[anchor=west,font=\fontsize{6.7}{7.8}\selectfont\itshape,text=black!65]
  at (-8.40,-1.90)
  {One shared decoder evaluates all $H$ horizon steps in parallel; no temporal mixing, recurrence, or iterative inverse kinematics.};
\end{tikzpicture}
\else
\includegraphics{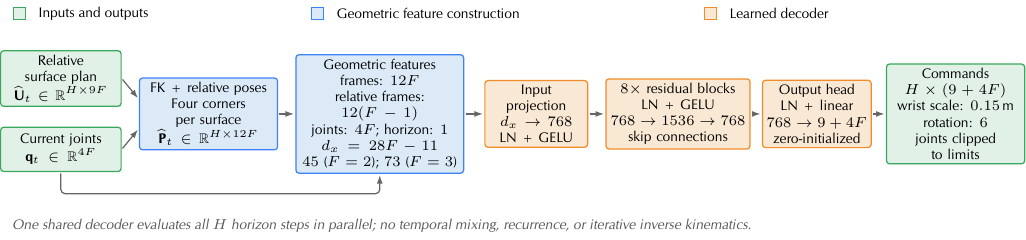}
\fi
\caption{\textbf{Contact-surface controller architecture.} The deterministic
geometric front end converts the planner's relative contact-surface trajectory
to ordered target corners and constructs absolute and reference-relative frame
features. A shared residual MLP then decodes every horizon step in one batched
forward pass. Green denotes inputs and outputs, blue denotes deterministic
geometric feature construction, and orange denotes the learned decoder.
Here $F$ is the number of controlled four-joint fingers, $d_x$ is the per-step
feature dimension, and $H=16$ is the trajectory horizon.}
\label{fig:supp_controller_architecture}
\end{figure*}

\subsubsection{Visual Augmentation}
Before frozen C-RADIOv3-B encoding, we cache one original and eight augmented
images per view. Four blend distinct MIL object or table
textures~\cite{finn2017oneshot} over the full image with original/texture weights
of 50/50, 60/40, 60/40, and 70/30, then apply photometric augmentation.
Four use photometric augmentation alone. Parameters are sampled independently
from Table~\ref{tab:supp_photo}. Figure~\ref{fig:supp_visual_augmentation}
shows the resulting candidates for a coin-picking demonstration.
Context images additionally use camera-extrinsic perturbations and
rendered-robot lighting and material variations before image-space augmentation.

\begin{table}[ht]
\centering
\footnotesize
\setlength{\tabcolsep}{4pt}
\begin{tabular*}{\columnwidth}{@{\extracolsep{\fill}}ll@{}}
\toprule
Parameter & Range \\
\midrule
Temperature $\tau$ & $[-0.40,0.40]$ \\
Tint $\eta$ & $[-0.12,0.12]$ \\
Exposure (EV) & $[-1,1]$ \\
Contrast & $[0.60,1.40]$ \\
Saturation & $[0.60,1.40]$ \\
Hue & $[-0.04,0.04]$ \\
Gamma & $[0.75,1.35]$ \\
Shading strength $a$ & $[-0.30,0.30]$ \\
Shading direction $\varphi$ & $[0,2\pi]$ \\
\bottomrule
\end{tabular*}
\par\vspace{4pt}
\caption{Uniform photometric-augmentation ranges; hue is measured in turns.
Temperature and tint set RGB gains to
$(e^{\tau-\eta/2},e^\eta,e^{-\tau-\eta/2})$.
Shading applies gain $1+a(x\cos\varphi+y\sin\varphi)/\sqrt{2}$,
where $x,y\in[-1,1]$ are normalized image coordinates.}
\label{tab:supp_photo}
\end{table}

Training samples uniformly from the nine candidates independently for each
view; validation uses originals. The no-visual-augmentation ablation uses only
originals while retaining the separate context-rendering selection.

\subsection{Contact-Surface Controller}
\label{sec:supp_controller}
This section details the controller in L3 of the main paper. Let $i$ index the
$F$ controlled fingers, $t$ be the current time, and $h$ be an offset within a
trajectory of length $H$.
The robot's current joints are $\mathbf q_t$, and
$\mathrm{FK}_i(\mathbf q_t)$ is contact surface $i$'s pose in the current wrist
frame. The predicted $\widehat{\Delta\mathbf T}_{i,t,h}$ gives its future pose
relative to its current contact-surface frame. Wrist poses are parameterized by
translation and the flattened first two rows of the rotation matrix.

\subsubsection{Geometric Inputs and Command Prediction}
The controller $C_\phi$ in the main paper takes the current joints and the full
contact-surface trajectory. It constructs geometric inputs and horizon coordinates
internally, then applies the shared network $c_\phi$ to all predicted steps in
parallel.
To express the trajectory in the current wrist frame, we compose each relative
contact-surface pose with its current forward-kinematics pose. We reconstruct the rotation from its
six-dimensional representation and compute each predicted corner as
\begin{equation}
\widehat{\mathbf p}_{i,\ell,t,h}
=\bigl[\mathrm{FK}_i(\mathbf q_t)
\widehat{\Delta\mathbf T}_{i,t,h}\bar{\mathbf v}_{i,\ell}\bigr]_{1:3}.
\label{eq:frame_to_contacts}
\end{equation}
Here $\bar{\mathbf v}_{i,\ell}$ is the homogeneous local corner defined in
Section~\ref{sec:supp_tracking}. The concatenated corners form
$\widehat{\mathbf P}_{t,h}\in\mathbb R^{12F}$. A deterministic residual MLP
predicts wrist and finger commands for each horizon step:
\begin{equation}
(\widehat{\Delta\mathbf X}_{t,h},\widehat{\mathbf q}_{t,h})
=c_\phi(\widehat{\mathbf P}_{t,h},\mathbf q_t,h/H),
\label{eq:surface_decoder}
\end{equation}
where $\widehat{\Delta\mathbf X}_{t,h}$ is the future wrist pose relative to the
current wrist, and $\widehat{\mathbf q}_{t,h}$ contains absolute joint targets.
The controller derives target-frame and thumb-relative-frame features from the
corners and combines them with normalized joints and $h/H$. Each frame feature
has 12 entries: three translation coordinates scaled by 10 and all nine
rotation-matrix entries, unlike the planner's nine-coordinate pose representation.
The output has $9+4F$ entries for the supported four-joint fingers. Joint targets
are clipped to their limits, although this does not guarantee that every
predicted multi-surface trajectory is kinematically reachable.

\subsubsection{Augmented Motion Supervision}
We augment wrist and joint trajectories using the endpoint perturbations in
Table~\ref{tab:supp_controller_settings}. For horizon step $h$, the smooth ramp
$\alpha_h=s_h^2(3-2s_h)$, with $s_h=h/H$, scales translation and joint offsets
and the angle of the left-multiplied rotation perturbation. After clipping joints
to their limits, forward kinematics supplies consistent contact-corner targets.
Following an action-regression warm-up, we minimize
\begin{equation}
\mathcal L_{\mathrm{ctrl}}=
\mathbb E\!\left[
\frac{1}{4F}\sum_{i,\ell}
\|100(\mathbf p^{\mathrm{FK}}_{i,\ell}(\widehat{\mathbf a})
-\mathbf p_{i,\ell})\|_2^2\right]
+\lambda_a\mathcal L_{\mathrm{act}},
\label{eq:decoder_loss}
\end{equation}
where $\widehat{\mathbf a}$ is the predicted wrist--joint action and
$\mathbf p_{i,\ell}$ is the target contact-surface corner. The factor 100 converts
positions from meters to centimeters before squaring the residual.
$\mathcal L_{\mathrm{act}}$ is normalized wrist--joint regression error, and
$\lambda_a=0.01$. The geometric term directly
penalizes realized contact-surface error, while action regression regularizes the
kinematic solution.

\subsubsection{Architecture and Optimization}
The residual MLP in Fig.~\ref{fig:supp_controller_architecture} is shared across
all horizon steps and evaluates them in parallel. Table~\ref{tab:supp_controller_settings}
lists its motion-augmentation and base optimization settings.

\begin{table}[t]
\centering
\footnotesize
\caption{Controller motion-augmentation and base optimization settings.}
\label{tab:supp_controller_settings}
\setlength{\tabcolsep}{3pt}
\renewcommand{\arraystretch}{1.08}
\begin{tabularx}{\columnwidth}{@{}P{0.35\columnwidth}Y@{}}
\toprule
Setting & Value \\
\midrule
\multicolumn{2}{@{}l}{\textit{Motion augmentation}} \\
Endpoint translation & Independent uniform offsets in $[-0.02,0.02]$\,m per axis \\
Endpoint rotation & Uniform angle in $[-0.15,0.15]$\,rad; uniformly sampled axis \\
Endpoint joint offsets & Gaussian; standard deviation 8\% of each joint's range \\
Application & Probability 0.9 per batch; independent perturbations per trajectory \\
\midrule
\multicolumn{2}{@{}l}{\textit{Base optimization}} \\
Batch sizes & Training 512; validation 128 \\
Warm-up & 1,000 action-only updates; LR $10^{-4}$ \\
Main objective & Geometric loss $+\,0.01\mathcal L_{\mathrm{act}}$; LR $3\times10^{-6}$ \\
AdamW & $\beta=(0.9,0.99)$; weight decay $10^{-6}$ \\
Stabilization & Gradient-norm clip 10; EMA decay 0.999 \\
Budget and validation & At most 20,000 updates; validate every 500 updates \\
Stopping target & $0.1$\,mm worst-task mean corner error \\
\bottomrule
\end{tabularx}
\end{table}


\subsection{Execution and Reuse}
\label{sec:supp_contact_execution}
The planner predicts $H=16$ targets at 10\,Hz, spanning 1.6\,s; execution uses an
eight-target chunk (0.8\,s at nominal speed). The 10\,Hz rate specifies trajectory
sampling, not replanning frequency. The controller remains frozen across tasks
that share the robot hand, controlled fingers, joint ordering, contact geometry,
and trajectory sampling; a different hand or finger set requires its own controller.

\subsection{Discussion on Next-State Action Targets}
\label{sec:supp_pose_force}
Following UMI~\cite{chi2024umi}, we use future demonstrated poses as action
targets, expressed relative to the current pose. These targets describe motion
rather than contact forces, but CIMI can still perform contact-rich manipulation
without force sensing. Three complementary mechanisms help explain this behavior.

\subsubsection{Compliant Contact}
With two friction-tape layers on the robot and one on the human interface,
the extra robot-side thickness can compress against the object at the same
nominal module pose. This tape deformation can produce contact loading without
targets inside the object.
UMI similarly uses passive fingertip compliance~\cite{chi2024umi}.

\subsubsection{Frictional Support}
The friction tape helps resist slip under the normal load supplied by robot
actuation. Its role is to support contact, not to measure or regulate force.

\subsubsection{Relative-Motion Feedback}
Replanning anchors predicted motion to the current configuration, allowing
visual and proprioceptive feedback to guide renewed corrections toward contact.
This is consistent with DexUMI's explanation of relative-action
robustness~\cite{xu2025dexumi}.

Future work could add post-contact tactile feedback~\cite{xu2025dexumi} or
augment pose demonstrations with force or wrench measurements for
force-informed actions~\cite{chen2025dexforce,zhang2025kinedex}.

\Needspace{4\baselineskip}
\subsection{Policy Camera Inputs and Preprocessing}
\label{sec:supp_policy_cameras}
Table~\ref{tab:supp_cameras} specifies the task-dependent camera selections.
\textbf{\texttt{CTX}} includes only the listed context views;
\textbf{\texttt{CTX+CC}} adds the listed contact views while retaining the same context inputs.
The dataset may store additional images for reconstruction or visualization;
these are not automatically supplied to the policy.

For the two-finger LEAP tasks, the human thumb and middle-finger contact modules
map to the robot thumb and index finger; the corresponding camera images follow
the same mapping. Bulb uses the human and robot thumb, index, and middle fingers.
Each selected stereo unit supplies one RGB policy view, using colors from its
rectified left image. Both stereo images are used for depth reconstruction;
they are not treated as two separate RGB policy inputs.

The deployment implementation rectifies the live stereo pair, reconstructs a
colored point cloud, and rasterizes it into the context image grid with
nearest-depth visibility. Stereo reconstruction uses an $800\times500$ grid,
and the policy context raster is $512\times320$. This preprocessing preserves
the rasterization used during training. The policy receives the resulting RGB
image, not depth or a point cloud. Human-to-robot viewpoint conversion and
robot-geometry overlay are performed on demonstration data offline.

\begin{table}[!ht]
\centering
\footnotesize
\setlength{\tabcolsep}{4pt}
\begin{tabular}{llll}
\toprule
Hand & Task & \textbf{\texttt{CTX}} & Added contact views \\
\midrule
LEAP & Cube & Wrist & Thumb, index \\
LEAP & Coin & Wrist & Thumb, index \\
LEAP & Letter & Wrist & Thumb, index \\
LEAP & Bulb & Wrist, web-space & Thumb, index, middle \\
Wuji & Cube, coin & Wrist & Thumb, middle \\
\bottomrule
\end{tabular}
\par\vspace{4pt}
\caption{Policy camera inputs. Finger names identify the controlled robot fingers.}
\label{tab:supp_cameras}
\end{table}

\subsection{Training and Inference Efficiency}
\label{sec:supp_learning_efficiency}
With cached features, the LEAP letter planner takes 24.3\,min for 63,000 updates
on an RTX 4090 D, including initialization, validation, and checkpointing.
Offline processing is reported separately in Table~\ref{tab:supp_processing_time}.

Table~\ref{tab:supp_inference_timing} uses the same GPU model: batch one, two
contact views, one wrist view, three flow steps, and 16 future targets.
RGB-to-command timing includes encoding and feature transfer, but excludes
capture, stereo reconstruction, rasterization, and robot communication.

\begin{table}[ht]
\centering
\footnotesize
\caption{Inference latency (ms). Both policy paths include the controller;
the timings are not additive.}
\label{tab:supp_inference_timing}
\setlength{\tabcolsep}{4pt}
\begin{tabular}{@{}lrr@{}}
\toprule
Timed path & Median & 95th percentile \\
\midrule
Prepared RGB to commands & 7.47 & 7.91 \\
Cached features to commands & 2.66 & 2.80 \\
Controller only (16 targets) & 0.84 & 0.98 \\
\bottomrule
\end{tabular}
\end{table}

\FloatBarrier
\balance

\renewcommand{\BIBdecl}{}
\begingroup

\endgroup

\else
  
\fi

\end{document}